\RequirePackage{fix-cm}

\documentclass[twocolumn]{svjour3}
\smartqed  

\usepackage{algorithm}
\usepackage{amsmath}
\usepackage{amsfonts}
\usepackage{rotating,color}
\usepackage{subfigure}
\usepackage{natbib}
\usepackage{multirow}
\usepackage{url}

\allowdisplaybreaks[3]

\usepackage{latexsym}
\usepackage{delarray}
\usepackage{graphicx}
\usepackage{psfrag}
\usepackage{lscape}
\usepackage{verbatim}

\def\boxit#1{\vbox{\hrule\hbox{\vrule\kern6pt\vbox{\kern6pt#1\kern6pt}\kern6pt\vrule}\hrule}}

\def\bd{{\boldsymbol d}}

\def\bs{{\boldsymbol s}}

\def\bT{{\boldsymbol T}}

\def\bV{{\boldsymbol V}}

\def\bx{{\boldsymbol{x}}}

\def\bX{{\boldsymbol X}}

\def\by{{\boldsymbol y}}

\def\bhbeta{{\widehat {\boldsymbol{\beta}}}}

\def\bzero{{\boldsymbol 0}}

\def\argmin{\mathop{\rm argmin}}
\def\real{\mathop{{\rm I}\kern-.2em\hbox{\rm R}}\nolimits}
\def\diag{\mbox{diag}}

\def\sgn{\hbox{sgn}}

\def\hbeta{\hat{\beta}}
\def\htheta{\hat{\theta}}

\def\bzero{{\boldsymbol 0}}

\def\bx{{\boldsymbol x}}
\def\bX{{\boldsymbol X}}

\def\bbeta{\boldsymbol \beta}

\def\lam{\lambda}

\journalname{Statistics and Computing}

\begin{document}


\title{APPLE: Approximate Path for Penalized Likelihood Estimators}

\author{Yi Yu         \and
        Yang Feng 
}


\institute{Y. Yu \at
              School of Mathematical Sciences, Fudan University, Shanghai, China, 200433 \\
              \email{yuyi@fudan.edu.cn}
\\
           \and
           Y. Feng \at
              Department of Statistics, Columbia University, New York, NY, U.S. 10027  \\
              \email{yangfeng@stat.columbia.edu}
}

\date{Received: date / Accepted: date}

\maketitle

\begin{abstract}
In high-dimensional data analysis, penalized likelihood estimators are shown to provide superior results in both variable selection and parameter estimation.  A new algorithm, APPLE, is proposed for calculating the Approximate Path for Penalized Likelihood Estimators. Both convex penalties (such as LASSO) and folded concave penalties (such as MCP) are considered. APPLE efficiently computes the solution path for the penalized likelihood estimator using a hybrid of the modified predictor-corrector method and the coordinate-descent algorithm. APPLE is compared with several well-known packages via simulation and analysis of two gene expression data sets.

\end{abstract}

\section{Introduction}
Variable selection is a vital tool in statistical analysis of high-dimensional data. Typically, a large number of potential predictors are included during the first stage of modeling, in order to avoid missing important links between a predictor and the outcome. This practice has become more popular in recent years for two primary reasons. First, in many recently promising fields, such as bioinformatics, genetics and finance, more and more high-throughput and high-dimensional data are being generated. Secondly, low cost and easy implementation for data collection and storage have made problems for which the number of variables is large, in comparison to the sample size, possible to be handled.

In order to provide more representative and reasonable applications of models in a mathematical framework, we often seek a smaller subset of important variables. The first attempt to variable selection was the $\ell_0$-type regularization methods, including AIC \citep{Akaike1973}, $C_p$ \citep{Mallows1973} and BIC \citep{Schwarz1978}, which work well in low-dimensional cases. In addition, they also exhibit good sampling properties
\citep{Barron.Birge.ea.1999}. However, searching all the possible subsets can be unstable \citep{Breiman1996}, and in high-dimensional settings, the
combinatorial problem has NP-complexity, which is
computationally prohibitive. As a result, numerous attempts have been made to modify the
$\ell_0$-type regularization to reduce the computational burden.   The most popular penalized regression method is LASSO \citep{Tibshirani1996} or equivalently Basis Pursuit \citep{ChenDonoho1994}. Being a convex penalty, it is computationally convenient, but lacks the oracle property and shrinks estimators regardless of importance. Hence, some folded concave penalties have been proposed in order to yield better performance, such as SCAD \citep{FanLi2001} and MCP \citep{Zhang2010}. We refer to \cite{FanXueZou} for the detailed definition of folded concave penalties. Also, for generalized linear models (GLM), penalized likelihood methods have been studied for high-dimensional variable selection, for example in \cite{FanLi2001} and \cite{vandeGeer2008}. We refer to  \citet{FanLv2010} for a review of variable selection in high-dimensionality.

Throughout the paper, we assume we have i.i.d. observations $(\bx_i, y_i), i=1,\cdots, n,$ where $\bx_i$ is a $p$-dimensional predictor and $y_i$ is the response. We further assume the conditional distribution of $y$ given $\bx$ belongs to an exponential family with canonical link, that is, it has the following density function
\begin{align}\label{eq::glm}
 f(y;\bx , \bbeta) = c(y) \exp\left[\frac{y\theta-b(\theta))}{a(\phi)}\right],
\end{align}
where $\theta=\bx'\bbeta$ and $\phi\in (0,\infty)$ is the dispersion parameter.

In view of \eqref{eq::glm}, the log-likelihood of the sample is given, up to an affine transformation, by
\begin{align*}
   \ell(\by; \bbeta)= n^{-1}\sum_{i=1}^n[y_i\theta_i-b(\theta_i)].
\end{align*}
Here, we are interested in estimating the $p$-dimensional vector $\bbeta$, and the penalized likelihood estimator is defined as
\begin{align*}
\bhbeta(\lambda)=\argmin_{\bbeta}\{-\ell(\by; \bbeta)+p_{\lambda}(\bbeta)\},
\end{align*}
where $p_{\lambda}(\cdot)$ is the LASSO or folded concave penalty function and $\lambda>0$ is the regularization parameter.

Developing an efficient algorithm for calculating the solution path of the coefficient vector $\bhbeta(\lambda)$, as $\lambda$ varies along a possible set of values, is very desirable. There is a vast literature on calculating such a path for penalized linear regression. For the convex penalty LASSO, least angle regression (LARS) \citep{EfronHJT2007}, or homotopy \citep{Osborne2000} are efficient methods for computing the entire path of LASSO solutions in the linear regression case. For folded concave penalties including SCAD and MCP, \cite{FanLi2001} used the local quadratic approximation (LQA); \cite{ZouLi2008} proposed the local linear approximation (LLA), which makes a local linear approximation to the penalty function, thereby yielding an objective function that can be optimized by using the LARS algorithm. \cite{Zhang2010} proposed the penalized linear unbiased selection (PLUS),  which is designed for the linear regression penalized by quadratic spline penalties, including LASSO, SCAD and MCP. More recently, coordinate descent methods have received considerable attention in high-dimensional settings, including \cite{Fu1998}, \cite{Shevade2003}, \cite{Krishnapuram2005}, \cite{Genkin2007}, \cite{FreidmanHHHT2007}, \cite{WuLange2008}, among others. Other work on penalized linear regression includes \cite{Hastie2004}, \cite{Daubechies2004}, \cite{Kim2007} and \cite{WeiZhu2012}.

Different from linear regression, derivatives of the log-likelihood in GLM are changing with respect to the regularization parameter $\lambda$. There has been major research on calculating the solution path for penalized likelihood estimators in the GLM setting. \cite{ParkHastie2007} proposed the \emph{glmpath} algorithm. They considered the solution $\bhbeta$ as a function of $\lambda$, and used a linear approximation of this function to update the estimator $\bhbeta$. They selected the step length in decreasing $\lambda$ by using an approximate smallest length that will change the active set of variables. \cite{YuanZou2009} approximated the loss function by a quadratic spline and showed a generalized LARS algorithm is suitable for solution path computation. It is worth to point out that their method can also be extended to more general regularization framework, including a generalization of the elastic net and a new method that effectively exploits the so-called “support vectors” in kernel logistic regression. \cite{FriedmanHT2010} proposed a coordinate descent algorithm for penalized GLM, in which they quadratically approximate the log-likelihood function and  sequentially solve the resulting penalized weighted least squares problem on a grid of $\lambda$ values.
\cite{Wu2010} proposed an ordinary differential equation-based solution path algorithm, which used quasi-likelihood instead of likelihood models, in order to use LARS more straightforwardly. However, all the numerical results in \cite{Wu2010} are based on the small $p$ large $n$ setting.
\cite{BrehenyHuang2011} adopted a coordinate descent algorithm for MCP and SCAD penalized GLM. Like \cite{FriedmanHT2010}, they used a quadratic approximation to the log-likelihood part and then used coordinate descent to update the regression parameter estimator. For the MCP penalty \citep{Zhang2010}, the tuning parameter $\gamma$ is used to adjust the concavity of the penalty. The smaller $\gamma$ is, the more concave the penalty is, which means finding a global minimizer is more difficult; but on the other hand, it results in less biased estimators.
The tuning parameter $\gamma$ can be changed freely from $1+$ to $\infty$. In the GLM case, \cite{BrehenyHuang2011} proposed the adaptive rescaling, which  allows the range of the parameter $\gamma$ to be as wide as it can be for the linear regression case. Other related papers  include \cite{ZhuHastie2004}, \cite{Lee2006}, \cite{RossetZhu2007} and \cite{MeiervanderGeer2008}.

In this work, we propose a new path algorithm, the \emph{approximate path for penalized likelihood estimators} (APPLE), under the setting of high-dimensional GLM.
Different from linear regression, it is often difficult to get explicit solutions in GLM.
Taking accuracy and feasibility into account, instead of linear approximation of the corresponding change in $\bbeta$ with the decrease in $\lambda$, which is used by most of the previous work, we use quadratic approximation to get a warm-start in updating. Then targeting on the KKT conditions, we perform a correction by optimizing a convex problem. Inspired by the adaptive rescaling in \cite{BrehenyHuang2011}, we develop a modified concavity adaptation method for MCP when updating the solution, which is shown to have better performance when $\gamma$  is small. In this paper, not only path algorithms for LASSO penalized GLM are derived, but also path algorithms for folded concave penalized GLM, which have appeared in few of the previous work. Here we mainly focus on MCP as an example of folded concave penalty, but it can be easily extended to other quadratic spline penalty functions.

For LASSO, we detect the  active set through the KKT conditions like most of the previous work. However, for some folded concave penalties, such as MCP, by fixing $\lambda$ and the concavity parameter $\gamma$, the value of the derivative of the penalty function decreases towards zero as the absolute value of the estimator increases towards $\lam\gamma$. We introduce a modified active set detection method, which has not appeared in any of the previous work.


The rest of the paper is organized as follows. In Sections \ref{sec-apple-lasso} and \ref{sec-apple-mcp}, we introduce the path algorithm APPLE for the LASSO and MCP penalties, respectively. We conduct simulation studies in Section \ref{sec-simulation} and  two real data examples are presented in Section \ref{sec-application}. A short summary is given in Section \ref{sec-summary}, while the technical details for logistic regression and Poisson regression are presented in the Appendix.

\section{APPLE with LASSO Penalty}\label{sec-apple-lasso}

LASSO \citep{Tibshirani1996} is a popular method for regression that uses an $\ell_1$-penalty to achieve simultaneous variable selection and parameter estimation. The idea has been broadly applied in GLM, where the problem is to minimize a convex function. In this section, we describe the details of the APPLE algorithm for LASSO penalized GLM.

\subsection{Problem Setup}

Let $\{(\bx_i,~y_i),~i=1,\cdots, n\}$ be $n$ i.i.d. pairs of $p$ predictors and a response as described in the introduction. By adding an additional column of $\mathbf{1}$'s to the design matrix $\bX$, the intercept $\beta_0$  is absorbed into the coefficient vector $\bbeta$. We are interested in finding the maximum likelihood solution for $\bbeta = (\beta_0, \beta_1, \cdots, \beta_p)'$, with a penalization on the size of the $\ell_1$-norm of the coefficients excluding the intercept. With a little abuse of notation, we denote $\|\bbeta\|_1 = \sum_{i=1}^p |\beta_j|$. Therefore, the optimization problem for a given $\lambda$ is reduced to finding $\bhbeta$, which minimizes the following:
\begin{align}\label{L}
& L_{\lambda}(\bbeta)=-\ell(\by;\bbeta)+\lambda \|\bbeta\|_1 \nonumber\\
& = -\frac{1}{n}\sum_{i=1}^n \{y_i \theta(\bbeta)_i-b(\theta(\bbeta)_i)\} +\lambda \|\bbeta\|_1.
\end{align}

As is common in GLM, the function $b(\theta)$ is implicitly assumed to be twice continuously differentiable with $b''(\theta)$ always positive.
It is straightforward to check that $L_{\lambda}(\cdot)$ is a convex function. Therefore, for a given $\lambda$, the unique minimizer $\bhbeta(\lam)$ is the solution to the KKT conditions, which are  given as follows.

\begin{align}\label{eq:KKT-LASSO}
\begin{cases}
\frac{\partial \ell}{\partial \beta_0}\bigg|_{\beta_0=\hbeta_0}=0, & \\
\frac{\partial \ell}{\partial \beta_j}\bigg|_{\beta_j=\hbeta_j}=\lambda \sgn(\hbeta_j) &  \mbox{for $j = 1,\cdots, p$, s.t. $\hbeta_j\neq 0$,}\\
\left|\frac{\partial \ell}{\partial \beta_j}\bigg|_{\beta_j=\hbeta_j}\right|\le\lambda  &  \mbox{for $j = 1, \cdots, p$, s.t. $\hbeta_j =0.$}
\end{cases}
\end{align}

\subsection{Grid of Penalty Parameter}

It is easy to notice from the KKT conditions that when
\begin{align*}
\lambda\ge\lambda_{\text{max}}=\max_{1\leq j\leq p}|\partial\ell/\partial\beta_j|_{\beta_j=0}|,
\end{align*}
$\hbeta_j=0$ for $1\leq j\leq p$.
As $\lambda$ decreases from $\lambda_{\max}$ to 0, $\bhbeta=\bhbeta(\lambda)$ changes from 0 (except for the intercept $\hbeta_0$) to the MLE solution. However, the full MLE solution has poor predictive performance and lacks the sparsity property because of the high-dimensionality. Here, following \cite{FriedmanHT2010} and \cite{BrehenyHuang2011}, we set the minimum value of $\lambda$ to be $\lambda_{\min}=\delta \lambda_{\max}$ and construct a sequence of $K$ values of $\lambda$ decreasing from $\lambda_{\max}$ to $\lambda_{\min}$ on the logarithm scale. We denote the sequence of $\lambda$ as $\lambda_k$, where $k=1,\cdots,K$. Typical values are $\delta = 0.01$ and $K=100$.

\subsection{Update}\label{subsec-lasso-update}
From the KKT conditions, we can see the relationship between $|\partial \ell/\partial \beta_j|$ and $\lambda$ determines whether the variable $\beta_j$ is activated or not. For  $\lambda_k$, we define the active set $A_k$ as follows,
\begin{align}\label{eq::lasso-active-set}
A_k=\left\{1\leq j\leq p:~ \left|\frac{\partial \ell}{\partial \beta_j}\right |\ge \lambda_k\right\} \cup \{0\},
\end{align}
and the step size as $\Delta_k=\lambda_{k+1}-\lambda_k$.
For a given $\lambda_k$ and active set $A_k$, we update the active coordinates together using the quadratic approximation,
\begin{align}\label{beta-approx}
\bhbeta^{(k+1,0)}_{A_k}=\bhbeta^{(k)}_{A_k}+\bs^{(k)}\cdot \Delta_k+\frac{1}{2}\bd^{(k)}\cdot \Delta_k^2,
\end{align}
where $\bs^{(k)}$ and $\bd^{(k)}$ are the first and second derivatives of $\bhbeta^{(k)}$ with respect to $\lambda$, respectively, which are derived using the chain rule, i.e.
\begin{align*}
& \left(\frac{\partial \ell}{\partial \bbeta_{A_k}}\right)\bigg|_{\bbeta = \bhbeta^{(k)}} = \lambda \sgn(\bhbeta^{(k)}_{A_k}) \\
\Rightarrow & \left(\frac{\partial^2 \ell}{\partial \bbeta_{A_k}\partial \bbeta^T_{A_k}}\cdot \frac{\partial \bbeta_{A_k}}{\partial \lambda}\right)\bigg|_{\bbeta = \bhbeta^{(k)}} = \sgn(\bhbeta^{(k)}_{A_k}) \\
\Rightarrow & \bs^{(k)} = \left(\frac{\partial^2 \ell}{\partial \bbeta_{A_k}\partial \bbeta^T_{A_k}}\right)^{-1}\bigg|_{\bbeta = \bhbeta^{(k)}}\cdot \sgn(\bhbeta^{(k)}_{A_k}),
\end{align*}
and,
\begin{align*}
\bd^{(k)} = \partial \left[\left(\frac{\partial^2 \ell}{\partial \bbeta_{A_k}\partial \bbeta^T_{A_k}}\right)^{-1}\cdot \sgn(\bhbeta^{(k)}_{A_k})\right] /\partial \lambda |_{\bbeta = \bhbeta^{(k)}}.
\end{align*}

 The explicit formula for calculating $\bs^{(k)}$ and $\bd^{(k)}$ are presented in the appendix for both logistic regression and Poisson regression. Since the intercept is not penalized, the first coordinates of $\bs^{(k)}$ and $\bd^{(k)}$ are both 0.
Different from \cite{ParkHastie2007}, where linear approximation was used, the quadratic approximation (\ref{beta-approx}) is more accurate and computationally efficient as a warm-start. Keep in mind, that here, the coefficients for the variables outside of $A_k$ are set to be 0 in $\bhbeta^{(k+1,0)}$. Additionally, the approximation (\ref{beta-approx}) will typically cause a small
deviation from the KKT conditions, which makes the following correction step necessary, in order to get the exact solution $\bhbeta^{(k+1)}$ at the current $\lambda_{k+1}$.

Here, we adopt two different correction methods depending on the current model size. To be more precise,  at step $k$, we check the following inequality,
\begin{align}\label{eq::NR-CD}
\# \{j: \bhbeta^{(k+1,0)}_j \neq 0\} \leq c \sqrt{n},
\end{align}
where $c$ is a user-specified constant. We set $c=1$ in all our numerical examples.  If \eqref{eq::NR-CD} holds (i.e., the current solution is relatively sparse compared with the sample size), we use a Newton-Raphson correction, otherwise, a coordinate descent correction is applied. When the correction method is stopped by a convergence check, the last $\bhbeta^{(k+1,j)}$ is denoted as $\bhbeta^{(k+1)}$.

\subsubsection{Newton-Raphson Correction}

Given the current solution $\bhbeta^{(k+1,0)}$, we  use the following Newton-Raphson method to correct the estimate until convergence,
\[
\bhbeta^{(k,j+1)}_{A_k}=\bhbeta^{(k,j)}_{A_k}-\left(\frac{\partial^2 L^{(k)}}{\partial \bbeta_{A_k}\bbeta^T_{A_k}}\right)^{-1} \left(\frac{\partial L^{(k)}}{\partial \bbeta_{A_k}}\right).
\]
Here, all the active variables are corrected together, which is different from coordinate descent method used in \cite{FriedmanHT2010} and \cite{BrehenyHuang2011}. We notice in our simulation studies that when \eqref{eq::NR-CD} holds, the Newton-Raphson type correction tends to be much faster than the coordinate descent correction method.

\subsubsection{Coordinate Descent Correction}

When \eqref{eq::NR-CD} does not hold (i.e., the number of active variables is relatively large), the Newton-Raphson method involves inverting a big matrix  $(\partial^2 L/\partial \bbeta^2)$, which may become ill-conditioned and cause stability issues in the iteration. Therefore, under this scenario, the more stable coordinate descent method is applied. In the coordinate descent algorithm, we fix all coefficients except $\beta_j$, and minimize (\ref{L})  for the current $\lam$ by updating $\beta_j$. The process is repeated for $j=0,\cdots,p$. After sweeping through all coordinates, we compare the new solution with $\bhbeta^{(k+1,0)}$. If they are sufficiently close, we have reached convergence; otherwise, the sweeping process is repeated until the two most recent estimators are close enough.  What makes the coordinate descent algorithm particularly attractive is that there is an explicit formula for each coordinate update. The details for the coordinate descent algorithm may be found in \cite{FriedmanHT2010} and \cite{BrehenyHuang2011}.

\subsection{Stopping Rules}\label{subsec-lasso-stop}

Following the updating process, we will obtain a solution path $\bhbeta^{(k)}$ for $k=1,\cdots,K$. However, from our simulation results, we notice that in most cases, the solutions near the end of the path  involve too many spurious variables. Therefore, the following two stopping rules are proposed to further speed up the path calculation process.
 \begin{enumerate}
   \item[(a)] (Model saturation detection). The first rule is designed to terminate the path algorithm if the fitting value is too extreme. For example,  in logistic regression, if the current estimated probability $\hat{p}_i = \exp(\bx_i'\bhbeta^{(k)})/(1+\exp(\bx_i'\bhbeta^{(k)}))$ satisfies
   \begin{align*}
   \max_{i=1,\cdots,n}\hat{p}_i>1-\epsilon \text{\quad or \quad } \min_{i=1,\cdots,n}\hat{p}_i<\epsilon,
   \end{align*}
   where $\epsilon$ is a predefined positive constant, we terminate the algorithm.
   \item[(b)] (A pre-specified maximum size of the model). In some real applications, the practitioner has an upper bound on the size of the model for various reasons. For example, in the optimal  portfolio allocation problem, one common restriction is the control of the transaction costs, which in turn puts a restriction on the maximum number of selected stocks.  In order to avoid missing the important variables, we usually set the upper limit significantly larger than the model size we need.
 \end{enumerate}
Although early stopping is performed following these two rules, the optimal solution always occurs before the stopping point in our numerical experience.

\subsection{Summary of the Algorithm}

\begin{enumerate}
\item[S1.] Define the grid of penalty parameters $\lambda$  as $\{\lambda_1,\cdots,\lambda_K\}$, where $\lambda_1=\lambda_{\max}$, $\lambda_K=\lambda_{\min} = \delta \lambda_{\max}$, and the remaining ones decrease on the logarithm scale. Set $k=1$ and the initial estimate $\bhbeta^{(1)}=\bzero$.

\item [S2.] Calculate the active set by $A_k=\left\{j:~ \left|\partial \ell/\partial \beta_j\right|\ge \lambda_{k} \right\} \cup \{0\}$.  Denote $\Delta_{k} = \lambda_{k+1} - \lambda_k$. The approximate solution is given by
\[
\bhbeta^{(k+1,0)}_{A_k}=\bhbeta^{(k)}_{A_k}+\bs^{(k)}\cdot \Delta_k+\frac{1}{2}\bd^{(k)}\cdot \Delta_k^2.
\]

\item [S3.] Correct the current solution towards the KKT conditions. If \eqref{eq::NR-CD} holds, we use the Newton-Raphson procedure; otherwise, coordinate descent method is adopted.

\item [S4.] Check  the two stopping rules, if at least one is satisfied, stop the algorithm; otherwise, set $k=k+1$ and repeat S2-S4 until $k=K$.

\end{enumerate}

\subsection{Selection of Tuning Parameter}\label{subsec-lasso-tune}

The performance of  penalized likelihood estimators depends heavily on the choice of  tuning parameters, that is $\lambda$ in LASSO and $(\lambda, \gamma)$ in MCP. This is usually accomplished through cross-validation or by using some information criterion such as AIC, BIC.

Information criteria derived using asymptotic arguments for the classical regression models are usually problematic when applied to penalized regression problems where $p\gg n$. For high-dimensional GLM, in \cite{ChenChen2008}, Extended BIC (EBIC) was proposed by adding an extra penalty term on top of BIC. It is defined as, for $0\le \gamma\le 1$,
\begin{align*}
\text{EBIC}_{\gamma}(\bs)=-2\log L_n\{\hat{\theta}(\bs)\}+\nu(\bs)\log n+ 2\gamma \log \binom{p}{j},
\end{align*}
where $\bs$ is a subset of $\{1,\cdots,p\}$, $\theta(\bs)$ is the parameter $\theta$ with those components outside $\bs$ being set to 0 or some pre-specified values, $\htheta(\bs)$ is the maximum likelihood estimator of $\theta(\bs)$, and $\nu(\bs)$ is the number of components in $\bs$. In this paper, we investigate both cross-validation and EBIC.

\section{APPLE with MCP Penalty}\label{sec-apple-mcp}
Different from LASSO, MCP is a folded concave penalty which was proposed by \cite{Zhang2010}. The penalty is a quadratic spline defined on $[0,\infty)$ by
\begin{align}\label{MCP-penalty}
p_{\lambda, \gamma}(t)=\lambda\int_0^t (1-x/(\gamma\lambda))_+~dx,
\end{align}
where the parameter $\gamma>0$ measures the concavity of the penalty, and $\lam$ is the regularization parameter. The APPLE algorithm for the MCP penalized GLM is slightly different from what we proposed in Section 2 for LASSO. Due to the non-convexity of MCP penalty, in this section, we will only focus on the main differences from the LASSO case.

\subsection{Problem Setup}

For MCP penalized GLM, the corresponding target function is

\begin{align}
& L_{\lambda}(\bbeta)=-\ell(\by;\bbeta)+\lambda\sum_{j=1}^p \int_0^{|\beta_j|} (1-\frac{x}{\lambda\gamma})_+~dx\nonumber\\
& =-\frac{1}{n}\sum_{i=1}^n \{y_i \theta(\bbeta)_i-b(\theta(\bbeta)_i)\}+\lambda\sum_{j=1}^p \int_0^{|\beta_j|} (1-\frac{x}{\lambda\gamma})_+~dx.\label{eq:MCP-1}
\end{align}

As introduced in \cite{Zhang2010} and \cite{ZhangHuang2008}, the sparse Riesz condition (SRC)($c_*, c^*, q$) holds under some mild regularity conditions. As a result, in the low-dimensional manifolds with dimension smaller than $q$, the convexity of $-\ell(\by; \bbeta)$ can dominate the concavity of the penalty, which will lead to the convexity of the target function (\ref{eq:MCP-1}) even with the choice of folded concave penalty. Therefore, under the SRC, for estimator with sparsity smaller than $q$, the KKT conditions are still valid to obtain a global minimizer.
The  KKT conditions are given as follows,
\begin{align}\label{eq:KKT-MCP-2}
\begin{cases}
\frac{\partial \ell}{\partial \beta_0}\big|_{\beta_0=\hbeta_0}=0, & \\
\frac{\partial \ell}{\partial \beta_j}\big|_{\beta_j=\hbeta_j}=\lambda (1-\frac{|\hbeta_j|}{\lambda\gamma}) \sgn(\hbeta_j) &  \mbox{for $0<|\hbeta_j|<\lambda\gamma$,} \\
\frac{\partial \ell}{\partial \beta_j}\big|_{\beta_j=\hbeta_j}=0 &  \mbox{for $|\hbeta_j|\ge\lambda\gamma$,} \\
|\frac{\partial \ell}{\partial \beta_j}\big|_{\beta_j=\hbeta_j}|\le\lambda & \mbox{for $\hbeta_j=0$.}
\end{cases}
\end{align}

\subsection{Grid of Penalty Parameter}
The grid $\{\lam_1, \cdots, \lam_K\}$ of penalty parameters is identical to that in the LASSO case.

\subsection{Update}\label{subsec-mcp-update}
In the LASSO case, the effective penalty level is $\lambda$ for all variables. Therefore, from the KKT condition (\ref{eq:KKT-LASSO}), as long as a variable is activated, it stays in the active set as $\lambda$ decreases. But in the MCP case, for the same $\lambda$, the effective penalty level on each variable is different depending on the magnitude of the estimate, as shown in \eqref{eq:KKT-MCP-2}. In all the existing work, such as NCVREG package, this specific property of MCP was not fully exploited and the same active set detection method was used as that for LASSO penalty (see \eqref{eq::lasso-active-set} for details). Here we introduce a new active set detection method using the KKT conditions, that, to our best knowledge, has not appeared before for folded concave penalties in the literature. As will be shown later, the new detection method is more suitable for MCP with a more efficient calculation. For a given $\lambda_k$, we define the active set $A_k$ as
\begin{equation*}
A_k=\{A_{k-1}\cup N_k\}\setminus D_k,
\end{equation*}
where
\begin{align*}
N_k = \big\{ j\in \{1,\cdots, p\}\setminus A_{k-1}:~|\partial \ell/\partial\beta_j|\ge \lambda_k\big\},
\end{align*}
and
\begin{align*}
D_k = \big\{j\in A_{k-1}\cap A_{k-2}:~\sgn(\hbeta_j^{(k-1)})\sgn(\hbeta_j^{(k-2)})<0\big\}.
\end{align*}
This means, with decreasing  threshold $\lam_k$, a particular variable becomes active when it satisfies  the KKT condition \eqref{eq:KKT-MCP-2}. Then the  variable will stay activated until it crosses 0 (i.e., the index lies in $D_k$), which means the covariates of the estimators have different signs in two consequent steps. From our experience, variables which cross 0 at some point in the path are usually noise variables.   If this deleted variable satisfies the KKT condition \eqref{eq:KKT-MCP-2} along the path later, we re-activate it. With decreasing $\lambda$, the optimization problem (\ref{eq:MCP-1}) will no longer be convex at some point. Therefore, the proposed treatment of deleting variables which cross 0 at some point will make the path more stable.

In accordance with Section \ref{subsec-lasso-update}, the step size is defined as $\Delta_k=\lambda_{k+1}-\lambda_k$. For a given $\lambda_k$ and  active set $A_k$, we update the active covariates altogether by using the quadratic approximation,
\[
\bhbeta^{(k+1, 0)}_{A_k}=\bhbeta^{(k)}_{A_k}+\bs^{(k)}\cdot \Delta_k+\frac{1}{2}\bd^{(k)}\cdot \Delta^{2}_k,
\]
where $\bs^{(k)}$ and $\bd^{(k)}$ are the first and second derivatives of $\bhbeta^{(k)}$ with respect to $\lambda$, respectively. Since the intercept is not penalized, the first coordinates of $\bs^{(k)}$ and $\bd^{(k)}$ are both 0.

Now, we have $\bs^{(k)} = (0, \bs^{(k)'}_{-0})'$, where $\bs^{(k)}_{-0}$ is defined as
\begin{align}\label{eq:adapt}
\bs^{(k)}_{-0}=\bigg[\frac{1}{n} \bX_{A_k\backslash\{0\}}^{'}
\bV^{(k)} \bX_{A_k\backslash\{0\}}-
{\bf \Gamma}\bigg]^{-1}(-\sgn(\bhbeta^{(k)}_{A_k\backslash\{0\}})),
\end{align}
where $\bV^{(k)}$ is given by (\ref{T}) in the Appendix, ${\bf \Gamma}=\text{diag}\{1/\gamma, \cdots,1/\gamma\}$, and $\mbox{sgn}(\cdot)$ is the sign function of a vector.

In MCP \citep{Zhang2010}, the tuning parameter $\gamma$ is free to vary from $1+$ to $\infty$. For the derivative $\bs^{(k)}$ defined in \eqref{eq:adapt}, particularly in logistic regression, $\gamma$ has to be large enough in order to make the matrix
\begin{align*}
n^{-1} \bX_{A_k\backslash\{0\}}^{'} \bV^{(k)} \bX_{A_k\backslash\{0\}}-\Gamma
\end{align*}
invertible.
However, if $\gamma$ is too large, the MCP penalty (\ref{MCP-penalty}) is approximately equal to $\lambda|t|$, which is the same as the LASSO penalty. In that case, it is hard to find the advantages which MCP enjoys over LASSO. In \cite{BrehenyHuang2011}, adaptive rescaling was proposed to solve a similar issue. They replaced $p_{\lambda,\gamma} (|\beta_j|)$ with $p_{\lambda, \gamma}(|v_j\beta_j|)$, where $v_j$ is the $j$-th diagonal element of the Hessian matrix. 
Since they used a coordinate descent algorithm, updating coordinates one-at-a-time,  the rescaled updates are straightforward after this change. But in our algorithm,  all the active variables are updated together. Therefore, a new adaptation method is needed.

The adaptation we use is to replace $p_{\lambda,\gamma}(|\beta_j|)$ by $p_{\lambda, \gamma u^{(k)}_{\min}}(|u_{\text{min}}^{(k)}\beta_j|)$, where $u_{\text{min}}^{(k)}$ is the smallest eigenvalue of the matrix $\frac{1}{n}\bX'_{A_k\backslash\{0\}}\bV^{(k)}\bX_{A_k\backslash\{0\}}$. Then,
\[
\bs_{-0}^{(k)}=\big(\frac{1}{n} \bX'_{A_k\backslash\{0\}}\bV^{(k)}\bX_{A_k\backslash\{0\}}-u_{\text{min}}^{(k)}{\bf \Gamma}\big)^{-1}(-\sgn(\bhbeta^{(k)}_{A_k\backslash\{0\}})).
\]
Now, for any $\gamma >1$,
\[
\frac{1}{n} \bX'_{A_k\backslash\{0\}}\bV^{(k)}\bX_{A_k\backslash\{0\}}-u_{\text{min}}^{(k)}{\bf \Gamma} > 0.
\]
Therefore, the singularity problem in (\ref{eq:adapt}) is avoided for all $\gamma>1$.

The correction method we introduced in the LASSO case is based on the fact that the problem (\ref{L}) is convex. Here in MCP, although the original problem is not convex, in each step after the adaptation,  our problem is still convex in a low-dimensional manifold as long as $\gamma>1$. One important issue is that when calculating the first and second order derivatives in the Newton-Raphson correction, $u_{\text{min}}^{(k)}$ is also a function of $\bhbeta^{(k)}$ (see Appendix). To avoid computing implicit derivatives, we use the popular quadratic approximation method (e.g., \cite{McCullagh1989}) to the negative log-likelihood, which turns out to be very effective. Our new target function is
\begin{align*}
& L(\lambda)=\frac{1}{2n}({\bf \tilde{y}}-\bX\beta)'\bV({\bf \tilde{y}}-\bX\beta)\\
& +\lambda\sum_{j=1}^p \int_0^{|u_{\text{min}}\beta_j|}(1-\frac{x}{\lambda\gamma u_{\text{min}}})_+~dx,
\end{align*}
where ${\bf \tilde{y}}=\bX\beta+\bV^{-1}(\by-{\bf \pi})$. The detailed formulations are presented in the Appendix, (\ref{piV11})-(\ref{piV13}) for logistic case and (\ref{piV21})-(\ref{piV23}) for Poisson case.

The same sparsity criterion (\ref{eq::NR-CD}) is used, and the corresponding Newton-Raphson or coordinate descent method is applied.

\subsection{Stopping Rules}

Stopping rules are the same as the ones discussed in Section \ref{subsec-lasso-stop}. But different from the LASSO case, if a variable is activated in a certain step of the MCP procedure (\ref{eq:KKT-MCP-2}), it may turn inactive later, and even be activated again later on. So for the same data, we can consider making the upper limit (Section \ref{subsec-lasso-stop}, (b)) in MCP larger than that would be for  the LASSO case.

\subsection{Summary of the Algorithm}
The algorithm is the same as in LASSO, except S2 is replaced with S2' described by the following.
\begin{enumerate}
\item [S2']  Calculate the active set by  $A_k= \{ A_{k-1}\cup N_k \}\setminus D_k$, where
\begin{align*}
N_k=\left\{j\in\{1,\cdots,p\}\setminus A_{k-1}:~ \left|\partial \ell/\partial \beta_j\right|\ge \lambda_k \right\},
\end{align*}
and
\begin{align*}
& D_k=\{j\in A_{k-1}\cap A_{k-2}:\\
&\sgn(\hbeta_j^{(k-1)})\sgn(\hbeta_j^{(k-2)})<0\}.
\end{align*}

\end{enumerate}

\subsection{Selection of Tuning Parameter}

The selection methods are mainly the same as the ones discussed in Section \ref{subsec-lasso-tune}. But as an advantage MCP is shown to possess in our numerical results, the estimators will stay in their optimal value for a certain interval of regularization parameter $\lambda$. This bears some advantages in selecting the tuning parameters.

\section{Simulation Results}\label{sec-simulation}

In this section, we conduct simulation studies for comparing APPLE with the GLMNET package \citep{FriedmanHT2010} and NCVREG package \citep{BrehenyHuang2011} for LASSO and MCP penalties, respectively. Now we highlight the differences of APPLE, GLMNET, and NCVREG. First, APPLE uses vectorized update when the estimator is sparse enough, which is faster than the coordinate descent method used in both GLMNET and NCVREG packages. Second, GLMNET is only available for convex penalties, while APPLE can handle both convex and non-convex penalties. Third, to deal with non-convex penalties, APPLE uses a different $\gamma$ adaptation and active set detection methods from those of NCVREG.

Logistic and Poisson regression models are two  popular generalized linear models. For each model, we present results of LASSO/MCP penalized methods. For the LASSO penalty, we compare APPLE LASSO with the GLMNET package \citep{FriedmanHT2010}.
For MCP, we compare APPLE MCP with the NCVREG package \citep{BrehenyHuang2011}. Since NCVREG only applies to Gaussian and logistic models, no comparable results are presented for MCP penalized Poisson model. For each setting, we report results for different tuning parameter selection methods, including EBIC and $K$-fold CV. The critera include \emph{false positives} (FP), \emph{true positives} (TP), \emph{$\ell_1 \text{  loss} =\|\bhbeta - \bbeta^o\|_1$}, \emph{$\ell_2 \text{ loss} =\|\bhbeta - \bbeta^o\|_2$}. We also compare the computational cost with NCVREG for the MCP penalty case.

\subsection{Logistic Regression}

\begin{example}\label{BLE1}
We consider a logistic regression model with different dimension, sparsity level and correlation settings. (i) Covariate dimension $p=1000$, sample size $n=500$ and $d=3$, where $d$ is the number of nonzero elements in $\bbeta^o$. The first 5 dimensions of $\bbeta^o$ are $(3, 1.5, 0, 0, 2)$, while the rest are all zeros and $\beta_0=0$. The vector $\bx$ follows a multivariate normal distribution with zero mean and covariance between the $i$-th and $j$-th elements being $\rho^{|i-j|}$ with $\rho=0, 0.2, 0.5$ and $0.7$ in four different settings. The results are summarized in Table \ref{BET1}. (ii) Different dimension and different sparsity levels are considered. In particular, $(p, n, d) = (3000, 500, 3)$ and $(1000, 500, 24)$. And for both settings, we consider $\rho = 0$ and $\rho = 0.5$ with the results reported in Table \ref{BET2}. When $d = 24$, the first 56 dimensions are 8 repetitions of $(3, 1.5, 0, 0, 2, 0, 0)$. In each setting, 100 repetitions are performed. Part of the setup is borrowed from \cite{FanLi2001}.
\end{example}

\begin{figure*}
\centering
\subfigure[]{
\includegraphics[scale=0.2]{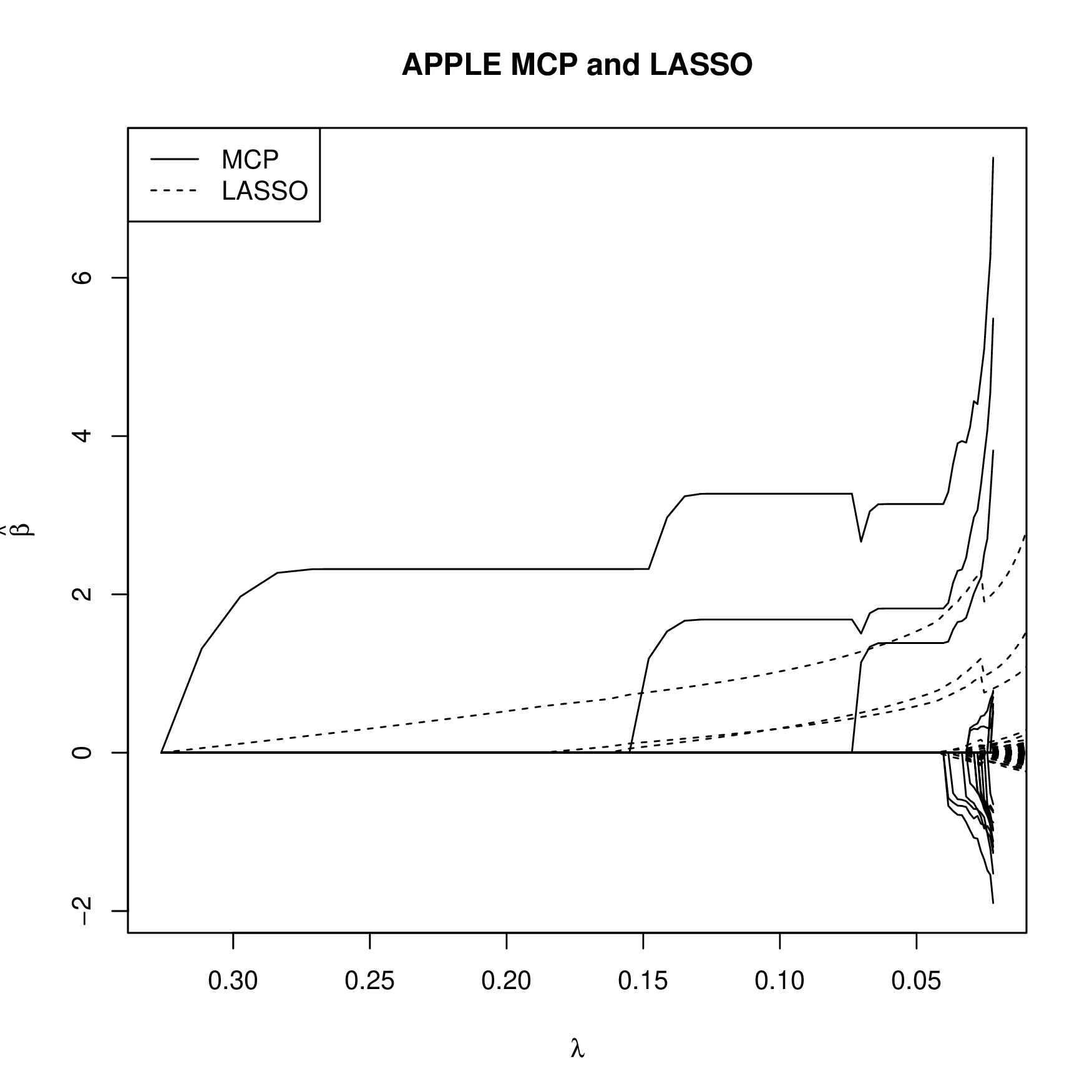}
\label{BL1}
}
\subfigure[]{
\includegraphics[scale=0.2]{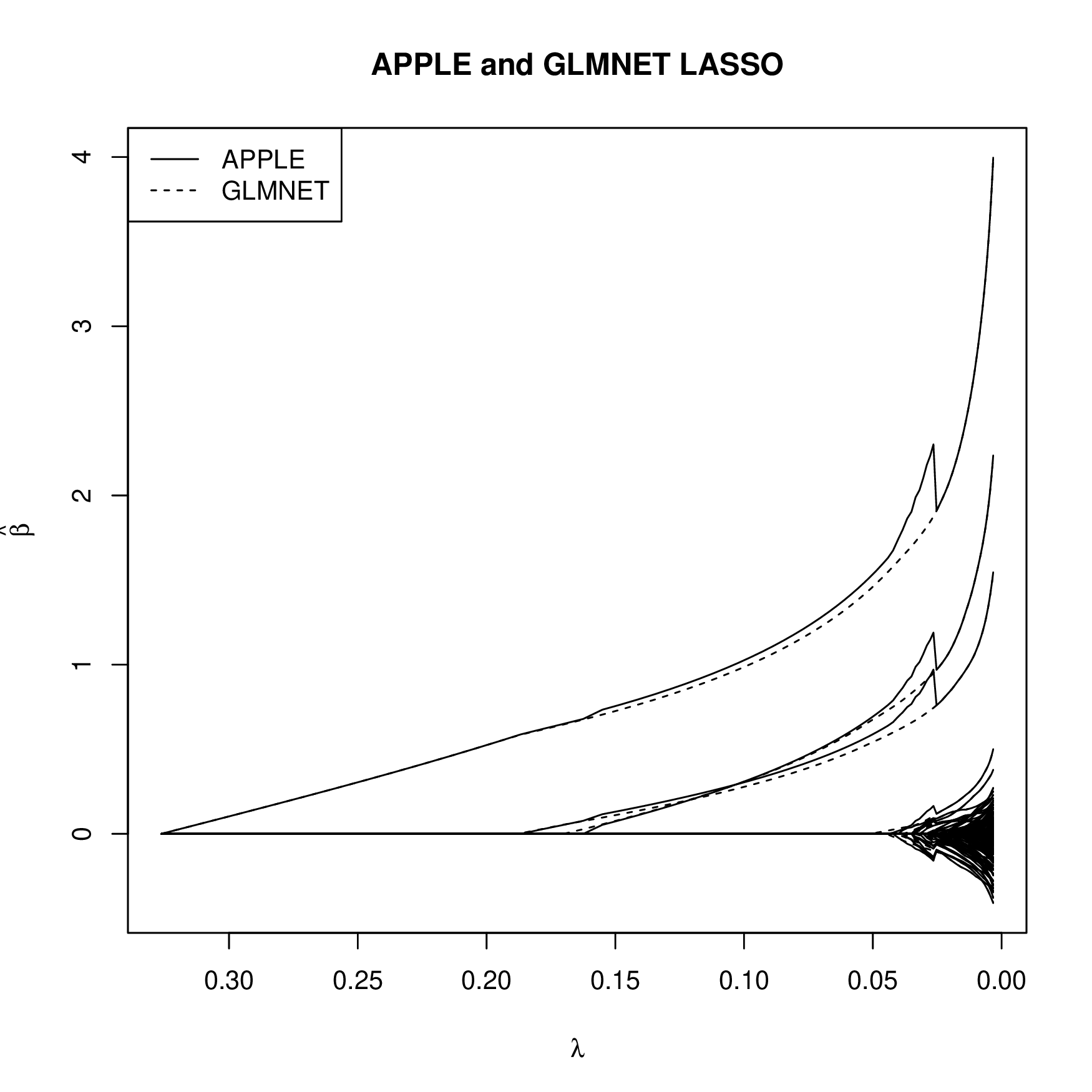}
\label{BL2}
}
\subfigure[]{
\includegraphics[scale=0.2]{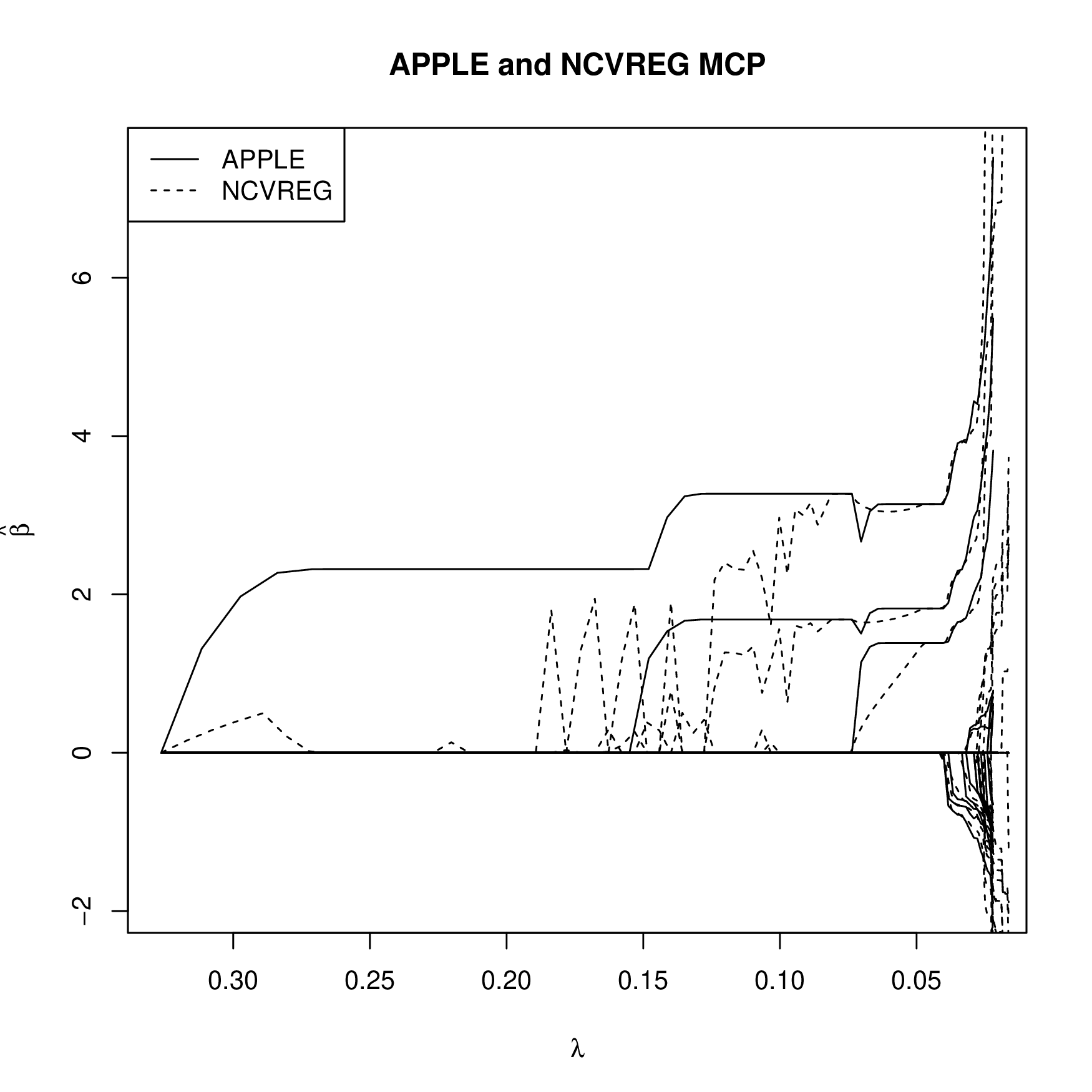}
\label{BL3}
}
\subfigure[]{
\includegraphics[scale=0.2]{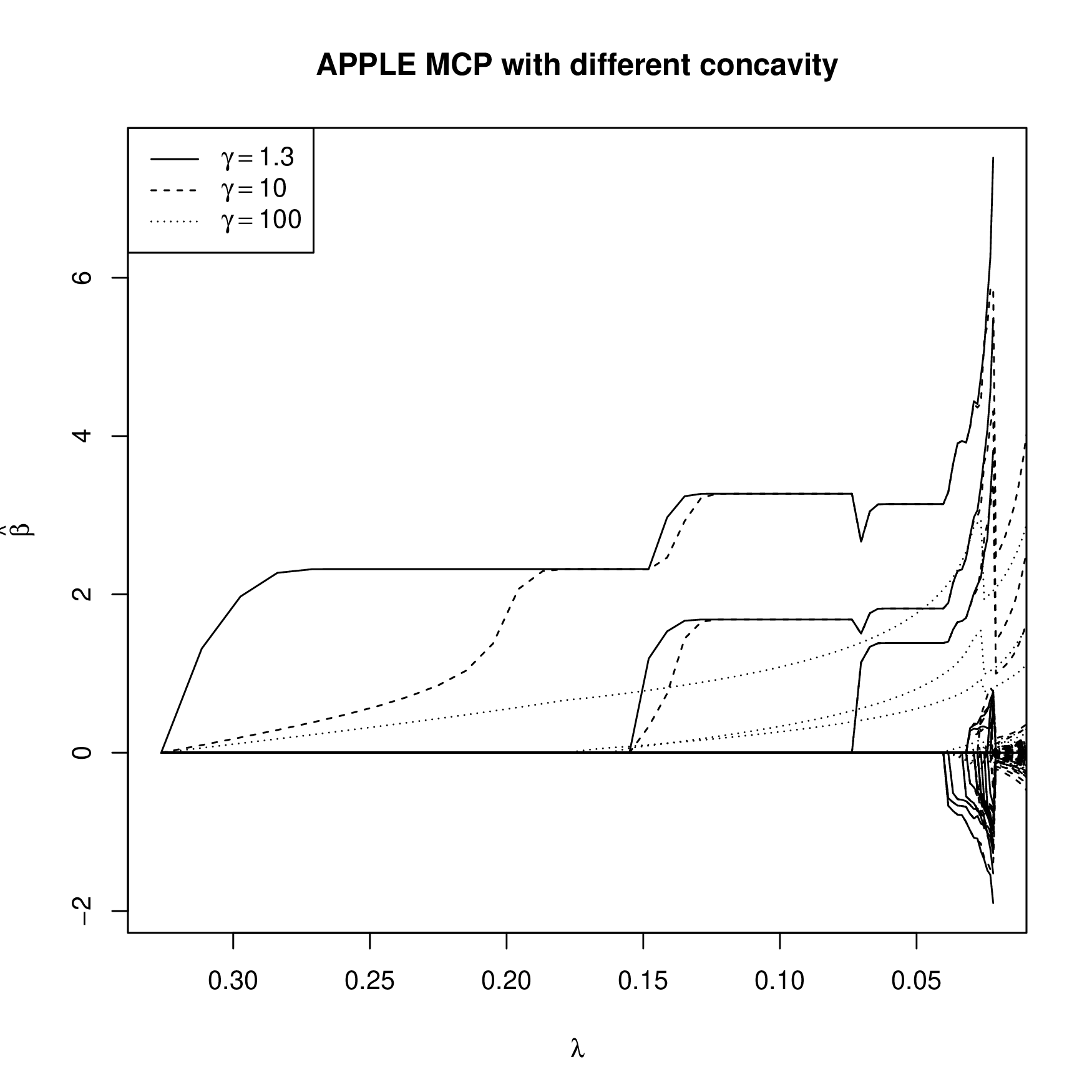}
\label{BL4}
}
\subfigure[]{
\includegraphics[scale=0.2]{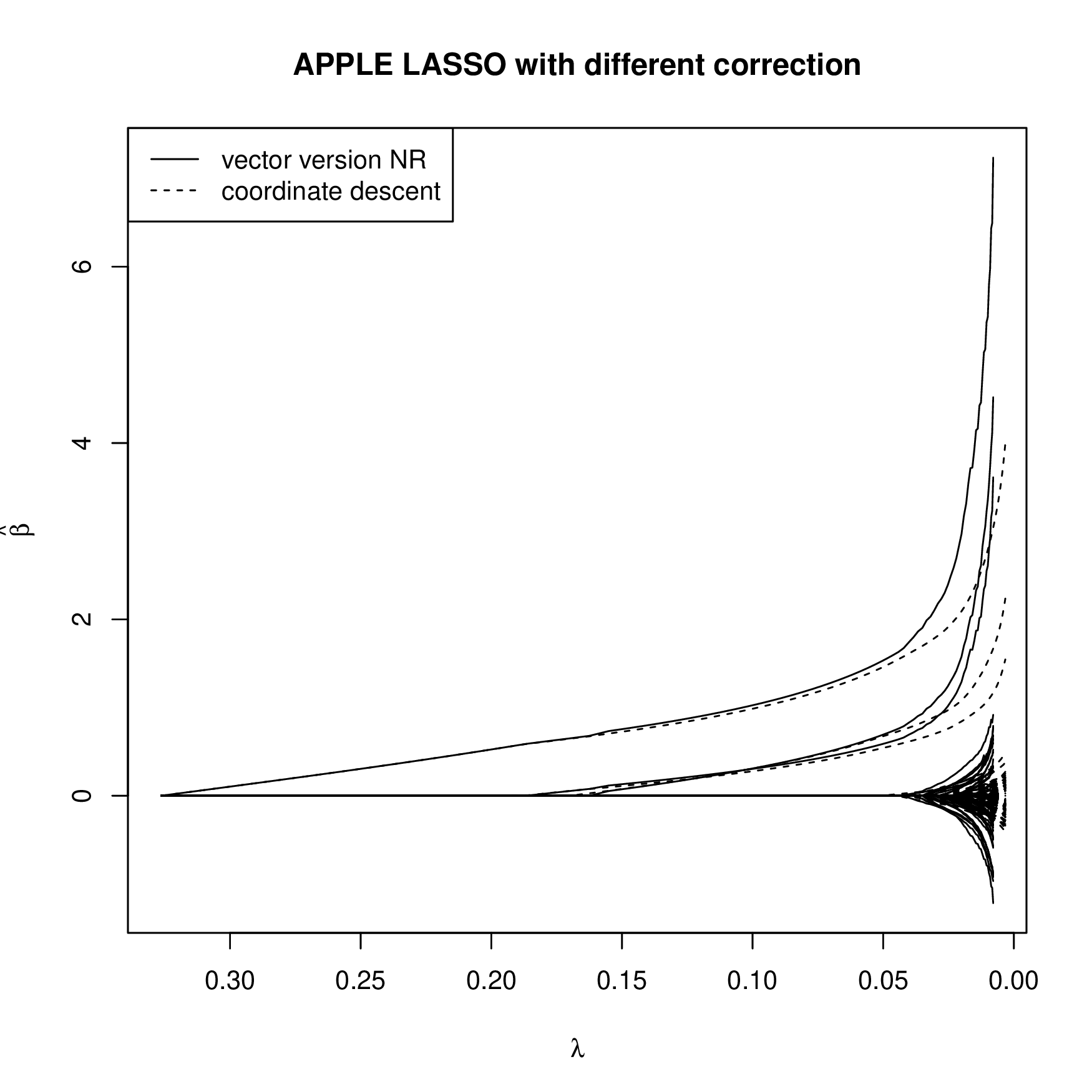}
\label{BL5}
}
\caption{Solution paths for logistic regression in Example \ref{BLE1}, where $p=1000$, $n=500$, $d=3$ and $\rho=0$. In (a), the solid lines and dotted lines are the solution paths for APPLE MCP and APPLE LASSO, respectively. In (b), the solid lines and dotted lines are the solution paths for APPLE LASSO and GLMNET LASSO, respectively. In (c), the solid lines and dotted lines are for APPLE and NCVREG MCP, respectively. In (d), the solid lines, dashed lines and dotted lines are solution paths of APPLE MCP with different $\gamma$ values. In (e), the solid lines and dotted lines are the solution paths for APPLE LASSO with different correction methods. For each panel and each type of lines, the important variables are selected in the same order. As $\lam$ becomes smaller, variables with index 1, 5 and 2 are selected one by one. When $\lam$ gets smaller than 0.05, noise variables are selected.}
\end{figure*}

\begin{table*}
\caption{Comparisons for APPLE with GLMNET and NCVREG for LASSO and MCP penalties, respectively, in Example \ref{BLE1}, where $p=1000$, $n=500$ and $d=3$. Design matrices with different correlation, and different selection criteria are presented. The medians of false positive (FP), true positive (TP), $\ell_1$ loss, and $\ell_2$ loss are reported over 100 repetitions, enclosed in parentheses are the corresponding standard errors. }
\label{BET1}
\centering
\begin{tabular}[h]{l | l l |  c c c c}
\hline\noalign{\smallskip}
Model & Package & Method &FP & TP & $\ell_1$ loss & $\ell_2$ loss \\[0.5ex]
\noalign{\smallskip}\hline\noalign{\smallskip}
LASSO & APPLE & EBIC & 0.30(0.54) & 3.00(0.00) & 3.64(0.28) & 4.44(0.70) \\
$\rho=0$ & & CV & 23.51(7.68) & 3.00(0.00) & 4.93(1.27) & 2.52(0.74) \\
& GLMNET & EBIC & 0.25(0.48) & 3.00(0.00) & 3.83(0.23) & 4.94(0.58) \\
& & CV & 48.44(26.02) & 3.00(0.00) & 4.93(1.27) & 2.52(0.74) \\
\noalign{\smallskip}\hline\noalign{\smallskip}
MCP & APPLE & EBIC & 0.03(0.17) & 3.00(0.00) & 0.68(0.38) & 0.21(0.25) \\
$\rho=0$, $\gamma=1.3$ & & CV & 0.26(0.03) & 3.00(0.00) & 0.86(0.96) & 0.34(0.67)\\
& NCVREG & EBIC & 0.65(0.45) & 3.00(0.00) & 2.50(0.31) & 1.51(0.42) \\
& & CV & 1.86(3.44) & 3.00(0.00) & 2.05(0.48) & 1.56(0.45)\\
\noalign{\smallskip}\hline\noalign{\smallskip}
MCP & APPLE & EBIC & 0.03(0.17) & 3.00(0.00) & 0.68(0.38) & 0.21(0.25) \\
$\rho=0$, $\gamma=3$ & & CV & 0.26(1.03) & 3.00(0.00) & 0.87(0.96) & 0.34(0.67) \\
& NCVREG & EBIC &  0.03(0.17) & 3.00(0.00) & 0.85(0.47) & 0.32(0.36)\\
& & CV &  2.65(5.37) & 3.00(0.00) & 1.04(0.78) & 0.30(0.32)\\
\noalign{\smallskip}\hline\noalign{\smallskip}
LASSO & APPLE & EBIC & 0.29(0.55) & 3.00(0.00) & 3.48(0.30) & 4.03(0.70)\\
$\rho=0.2$ & & CV & 19.96(11.60) & 3.00(0.00) & 3.70(1.53) & 2.31(0.61)\\
& GLMNET & EBIC & 0.23(0.48) & 3.00(0.00) & 3.74(0.26) & 4.71(0.67)\\
& & CV & 83.19(11.60) & 3.00(0.00) & 4.73(1.53) & 2.31(0.61)\\
\noalign{\smallskip}\hline\noalign{\smallskip}
MCP & APPLE & EBIC & 0.02(0.15) & 3.00(0.00) & 0.72(0.36) & 0.22(0.25) \\
$\rho=0.2$, $\gamma=1.3$ & & CV & 0.14(0.59) & 3.00(0.00) & 0.80(0.63) & 0.29(0.47)\\
& NCVREG & EBIC & 0.03(0.15) & 3.00(0.00) & 0.72(0.36) & 0.22(0.25)\\
& & CV & 0.22(0.73) & 3.00(0.00) & 0.82(0.55) & 0.27(0.34)\\
\noalign{\smallskip}\hline\noalign{\smallskip}
MCP & APPLE & EBIC & 0.02(0.15) & 3.00(0.00) & 0.72(0.36) & 0.22(0.25) \\
$\rho=0.2$, $\gamma=3$ & & CV & 0.14(0.59) & 3.00(0.00) & 0.80(0.63) & 0.29(0.47)\\
& NCVREG & EBIC & 0.02(0.14) & 3.00(0.00) & 0.72(0.37) & 0.25(0.22) \\
& & CV & 0.42(0.73) & 3.00(0.00) & 0.80(0.64) & 0.30(0.34)\\
\noalign{\smallskip}\hline\noalign{\smallskip}
LASSO &APPLE & EBIC & 0.10(0.30) & 3.00(0.00) & 3.63(0.27) & 4.44(0.70) \\
$\rho=0.5$ & & CV &  19.20(0.45) & 3.00(0.00) &  3.72(0.26)  & 2.24(0.64)\\
& GLMNET & EBIC & 0.10(0.30) & 3.00(0.00) & 3.78(0.22) & 4.84(0.58)\\
& & CV & 43.88(26.02) & 3.00(0.00) & 4.77(1.27) & 2.48(0.74)\\
\noalign{\smallskip}\hline\noalign{\smallskip}
MCP & APPLE & EBIC & 0.02(0.15) & 3.00(0.00) & 0.66(0.32) & 0.18(0.20) \\
$\rho=0.5$, $\gamma=1.3$ & & CV & 0.12(0.25) & 3.00(0.00) & 0.79(0.49) & 0.30(0.26) \\
& NCVREG & EBIC & 0.07(0.26) & 2.98(0.15) & 0.73(0.44) & 0.25(0.44) \\
& & CV & 3.33(2.67) & 3.00(0.00) & 2.84(0.42) & 2.77(0.85) \\
\noalign{\smallskip}\hline\noalign{\smallskip}
MCP & APPLE & EBIC & 0.01(0.10) & 3.00(0.00) & 0.72(0.34) & 0.22(0.22)\\
$\rho=0.5$, $\gamma=3$ & & CV & 0.14(0.41) & 3.00(0.10) & 0.87(0.60) & 0.34(0.47)\\
& NCVREG & EBIC & 0.01(0.11) & 3.00(0.00) & 1.62(0.54) & 0.96(0.55) \\
& & CV & 4.93(5.60) & 3.00(0.00) & 1.55(0.97) & 0.58(0.52)\\
\noalign{\smallskip}\hline\noalign{\smallskip}
LASSO & APPLE & EBIC & 0.44(0.61) & 3.00(0.00) & 3.52(0.24) & 4.15(0.60)\\
$\rho=0.7$ & & CV & 17.22(18.54) & 3.00(0.00) & 3.64(1.10) & 3.48(1.57)\\
& GLMNET & EBIC & 0.51(0.63) & 3.00(0.00) & 3.72(0.26) & 5.65(0.64)\\
& & CV & 45.77(11.26) & 3.00(0.00) & 4.76(1.62) & 3.88(1.73) \\
\noalign{\smallskip}\hline\noalign{\smallskip}
MCP & APPLE & EBIC & 0.16(0.49) & 2.87(0.34) & 1.19(0.76) & 0.72(0.95)\\
$\rho=0.7$, $\gamma=1.3$ & & CV & 0.65(1.02) & 2.79(0.41) & 1.68(1.07) & 1.18(1.20)\\
& NCVREG & EBIC & 0.26(0.62) & 2.81(0.39) & 1.56(0.78) & 1.09(1.01)\\
& & CV & 0.72(0.90) & 2.85(0.36) & 1.59(0.68) & 1.07(0.91)\\
\noalign{\smallskip}\hline\noalign{\smallskip}
MCP & APPLE & EBIC &  0.16(0.53) & 2.88(0.32) & 1.20(0.69) & 0.72(0.97)\\
$\rho=0.7$, $\gamma=3$ & & CV & 0.65(1.02) & 2.79(0.41) & 1.68(1.07) & 1.18(1.20)\\
& NCVREG & EBIC & 0.20(0.59) & 2.80(0.39) & 1.56(0.78) & 1.09(1.01)\\
& & CV &  0.70(0.95) & 2.89(0.36) & 1.56(0.77) & 1.17(0.95)\\
\noalign{\smallskip}\hline
\end{tabular}
\end{table*}

\begin{table*}
\caption{Comparisons for APPLE with GLMNET and NCVREG for LASSO and MCP penalties, respectively, in Example \ref{BLE1}, where $(p, n, d) = (3000, 500, 3)$ and $(1000, 500, 24)$. Design matrices with different correlation, and different selection criteria are presented. The medians of false positive (FP), true positive (TP), $\ell_1$ loss, and $\ell_2$ loss are reported over 100 repetitions, enclosed in parentheses are the corresponding standard errors. }
\label{BET2}
\centering
\begin{tabular}[h]{l | l l |  c c c c}
\hline\noalign{\smallskip}
Model & Package & Method &FP & TP & $\ell_1$ loss & $\ell_2$ loss \\[0.5ex]
\noalign{\smallskip}\hline\noalign{\smallskip}
LASSO & APPLE & EBIC & 0.06(0.09) & 3.00(0.00) & 3.73(0.67) & 4.70(0.84)\\
$p = 3000$ & & CV & 53.93(15.90) & 3.00(0.00) & 5.27(1.52) & 2.50(0.54) \\
$d = 3$ & GLMNET & EBIC & 0.07(0.05) & 3.00(0.00) & 4.00(1.09) & 5.43(1.10)\\
$\rho=0$& & CV &  118.03(20.83) & 3.00(0.00) & 5.79(1.46) & 3.94(0.39f)\\
\noalign{\smallskip}\hline\noalign{\smallskip}
MCP & APPLE & EBIC & 0.03(0.18) & 3.00(0.00) & 0.77(0.41) & 0.24(0.26) \\
$p = 3000$ & & CV & 0.23(0.25) & 3.00(0.00) & 0.78(0.39) & 0.23(0.22) \\
$d = 3$& NCVREG & EBIC & 0.03(0.10) & 3.00(0.00) & 0.78(0.40) & 0.25(0.22)\\
$\rho=0$ & & CV & 0.27(0.64) & 3.00(0.00) & 1.02(0.88) & 0.44(0.67)\\
\noalign{\smallskip}\hline\noalign{\smallskip}
LASSO & APPLE & EBIC & 0.19(0.43) & 3.00(0.00) & 3.61(0.26) & 4.40(0.62)\\
$p = 3000$ & & CV & 32.93(42.06) & 3.00(0.00)& 4.28(1.89) & 2.58(0.72)\\
$d = 3$ & GLMNET & EBIC & 0.21(0.44) & 3.00(0.00) & 3.83(0.26) & 4.98(0.65)\\
$\rho=0.5$& & CV & 190.17(13.96) & 3.00(0.00) & 3.83(0.25) & 4.99(0.65) \\
\noalign{\smallskip}\hline\noalign{\smallskip}
MCP & APPLE & EBIC &  0.16(0.54) & 3.00(0.00) & 1.03(0.68) & 0.50(0.64)\\
$p = 3000$ & & CV &  0.01(0.10) & 3.00(0.00) & 0.89(0.52) & 0.38(0.50)\\
$d = 3$ & NCVREG & EBIC & 0.30(0.66) & 3.00(0.00) & 1.03(0.54) & 0.53(0.49) \\
$\rho=0.5$& & CV & 0.02(0.15) & 3.00(0.00) & 0.94(0.53) & 0.41(0.51)\\
\noalign{\smallskip}\hline\noalign{\smallskip}
LASSO & APPLE & EBIC & 0.24(0.52) & 11.17(7.67) & 7.01(2.37) & 10.35(3.36) \\
$p = 1000$ & & CV & 124.51(18.00) & 23.90(0.29) & 6.85(0.99) & 8.52(2.36) \\
$d = 24$ & GLMNET & EBIC & 0.00(0.00) & 0.12(0.54) & 7.21(0.10) & 11.03(0.58) \\
$\rho=0$& & CV & 200.41(12.39) & 23.91(0.28) & 7.20(1.49) & 6.73(1.82) \\
\noalign{\smallskip}\hline\noalign{\smallskip}
MCP & APPLE & EBIC &  0.12(0.38) & 20.63(1.99) & 4.48(2.50) & 4.68(3.35)\\
$p = 1000$ & & CV &  0.09(0.29) & 23.25(1.42) & 4.16(2.32) & 4.17(2.86) \\
$d = 24$ & NCVREG & EBIC & 0.14(0.36) & 21.95(2.31) & 4.38(3.68) & 5.21(6.45)\\
$\rho=0$& & CV & 4.59(1.70) & 23.54(0.81) & 17.82(11.51) & 70.61(74.69)\\
\noalign{\smallskip}\hline\noalign{\smallskip}
LASSO & APPLE & EBIC & 0.90(0.98) & 14.67(5.69) & 6.94(1.80) & 10.16(2.91)\\
$p = 1000$ & & CV & 110.84(14.05) & 23.82(0.43) & 6.82(1.01) & 8.45(2.32)\\
$d = 24$ & GLMNET & EBIC & 0.00(0.00) & 0.88(3.31) & 7.20(0.61) & 11.01(3.03)\\
$\rho=0.5$& & CV &  167.15(10.24) & 23.80(0.44) & 7.06(2.12) & 7.00(2.94)\\
\noalign{\smallskip}\hline\noalign{\smallskip}
MCP & APPLE & EBIC &  0.90(0.97) & 15.78(2.12) & 5.51(2.17) & 6.54(3.51)\\
$p = 1000$ & & CV &  1.14(1.09) & 16.93(1.88) & 5.16(2.20) & 5.88(3.21)\\
$d = 24$ & NCVREG & EBIC & 0.69(0.83) & 15.61(2.34) & 5.37(2.70) & 6.41(4.02)\\
$\rho=0.5$& & CV &  6.95(3.35) & 18.29(1.99) & 15.87(11.20) & 58.76(53.40)\\
\noalign{\smallskip}\hline
\end{tabular}
\end{table*}

In Figure \ref{BL1}, we compare the solution paths of the APPLE algorithm for MCP and LASSO. We can see that the MCP path is less smooth than the LASSO path, but has intervals at which the estimators stay constant, and yields a sparser model. With the convergence stopping rules adopted in all simulations here, the corresponding solutions of the MCP path are sparse even near the end of the path. Actually, the size of active set does not exceed the square root of the sample size, which means that the Newton-Raphson correction is used throughout the whole path. Notice that there is a jump on the LASSO path, which is caused by the change of correction method. See \cite{Feng-2011} for the stability comparison of various penalty functions.
In Figure \ref{BL2}, the APPLE and GLMNET LASSO paths are illustrated. Before the change point, using the Newton-Raphson correction method, the APPLE path exhibits better estimation with a sparser model. After the change point, coordinate descent correction  is employed, which  makes the two paths identical.
In Figure \ref{BL3}, the APPLE and NCVREG MCP paths are compared given the same concavity parameter $\gamma=1.3$. APPLE paths are significantly smoother than NCVREG paths. Although both paths stay at the ``optimal" level, APPLE paths have a longer period, which makes the model selection task easier and leads to more stable estimation. 
In Figure \ref{BL4}, APPLE MCP paths with different concavity parameters ($\gamma=1.3, 3, 100$) are presented. This shows that as $\gamma$ gets larger, the ``flat" period of constant optimal magnitude gets shorter, and the APPLE MCP path eventually approaches the LASSO path  when $\gamma$ becomes sufficiently large.
In Figure \ref{BL5}, we show APPLE LASSO paths with different correction methods throughout the whole path. We can see that at the beginning of the path when $\lambda$ is large, the differences between these two correction methods are negligible. However, as $\lambda$ decreases, more variables are recruited and the Newton-Raphson method becomes unstable, making the coefficient estimates ``take-off'' more quickly compared with the coordinate descent method. Therefore, we recommend the hybrid approach of using the Newton-Raphson in the first part of the path, and later switch to coordinate descent when the number of active variables becomes large enough.

The FP, TP, $\ell_1$ loss and $\ell_2$ loss results for Example \ref{BLE1}(i) are summarized in Table \ref{BET1}.  When $\rho=0$,  comparing the APPLE LASSO and GLMNET, we see the results from EBIC are similar for these two methods. However, for the CV, APPLE tends to provide a model with smaller FP values while keeping TP the same.
When MCP is applied, similar observations can be found. 
Overall, comparing with the existing methods, APPLE does a better job than GLMNET and NCVREG in the LASSO and MCP cases, respectively. 
In addition, for the MCP penalty, APPLE provides a smoother path than NCVREG.

The corresponding results for Example \ref{BLE1}(ii) can be found in Table \ref{BET2}. When the dimension is increased to 3000 from 1000, the behaviors of APPLE applied to both LASSO and MCP cases are similar to those analyzed in Example \ref{BLE1}(i). Recall that in APPLE, we use two different correction methods when the size of the current active set changes. To study the robustness of this dynamic correction method, we consider the case $d = 24$, which implies $d > \sqrt{n}$, i.e., the true model size exceeds the square root of sample size. In this scenario, for each setting and each method, some important variables are missing, particularly for EBIC based LASSO estimators. We conjecture the reason to be the over-penalization of EBIC. Recall that the modification in EBIC is a prior imposed on all the possible subsets of a given sparsity level. As a result, when the size of active set is large, the penalty is too stringent for the variable selection purpose. Nevertheless, with the same method for choosing the tuning parameter, APPLE does a better job than GLMNET and NCVREG in the LASSO and MCP cases, respectively.

\begin{table*}
\caption{Comparison of the computational cost for the APPLE and NCVREG packages in different simulation settings. The medians of the computation time (in seconds) are reported, enclosed in parentheses are the corresponding standard errors. CPU: Intel(R) Xeon(R) L5420 @ 2.50GHz. }\label{time}
\centering
\begin{tabular}[h]{l l |  c c| c c}
\hline\noalign{\smallskip}
& $\gamma$ &\multicolumn{2}{c|}{$\rho=0$} &\multicolumn{2}{c}{$\rho=0.5$}\\[0.5ex]
& &NCVREG & APPLE & NCVREG & APPLE \\
\noalign{\smallskip}\hline\noalign{\smallskip}
$p=2^7$ & 1.3 & 1.33(0.41)& 0.13(0.07) & 1.42(0.52) & 0.14(0.06)\\
$n=50$ & 3& 0.22(0.28) & 0.06(0.04) & 0.15(0.32) & 0.07(0.02) \\
\noalign{\smallskip}\hline\noalign{\smallskip}
$p=2^8$ &1.3 & 4.09(1.17) &0.37(0.21)& 5.58(1.29) & 0.42(0.11) \\
$n=100$ & 3 &0.35(0.53) & 0.19(0.06) & 0.32(0.40) &0.27(0.10)\\
\noalign{\smallskip}\hline\noalign{\smallskip}
$p=2^9$ & 1.3 & 18.11(4.47) & 1.15(0.18) & 27.47(6.25) & 1.21(0.20)\\
$n=200$ & 3 & 1.58(0.77) & 0.86(0.13) & 1.32(0.55) & 1.08(0.14)\\
\noalign{\smallskip}\hline\noalign{\smallskip}
$p=2^{10}$ & 1.3 &123.53(22.87) &6.29(0.67)  & 186.95(33.00) &6.55(0.80)\\
$n=500$ & 3 & 6.55(1.49) &5.55(0.66)  & 9.01(4.42) & 6.03(0.65)\\
\noalign{\smallskip}\hline
\end{tabular}
\end{table*}

APPLE is an efficient algorithm for computing the solution path for penalized likelihood estimators, particularly for folded concave penalties. Table \ref{time} illustrates the median time required to fit the entire path and the corresponding standard errors of the NCVREG and APPLE algorithms. Here, we use the same setting as Example \ref{BLE1} except for different $p$ and $n$. It is clear that APPLE takes less time than NCVREG for the current example.

\subsection{Poisson Regression}\label{PR}


\begin{example}\label{PRE1}
We consider a Poisson regression model with different dimension, sparsity level and correlation settings. (i) As Example \ref{BLE1}(i), we set $(p,n,d)=(1000,500,3)$. The first 5 dimensions of $\bbeta^o$ are $(1.2, 0.6, 0, 0, 0.8)$, and the rest are all zeros and $\beta_0=0$. The vector $\bx$ follows a multivariate normal distribution with zero mean and covariance between the $i$-th and $j$-th elements being $\rho^{|i-j|}$ with $\rho=0, 0.2, 0.5$ and $0.7$ in four different settings. The results are in Table \ref{PET1}. (ii) As Example \ref{BLE1}(ii), different dimension and different sparsity levels are considered. We consider both $(p, n, d) = (3000, 500, 3)$ and  $(1000, 500, 24)$ with $\rho = 0$ and $\rho = 0.5$.  All the results are summarized in Table \ref{PET2}. When $d = 24$, the first 56 dimensions are 8 repetitions of $(1.2, 0.6, 0, 0, 0.8)$. In each setting, 100 repetitions are performed. Part of the setup is borrowed from \cite{ZouLi2008}.
\end{example}

\begin{figure*}
\centering
\subfigure[]{
\includegraphics[scale=0.30]{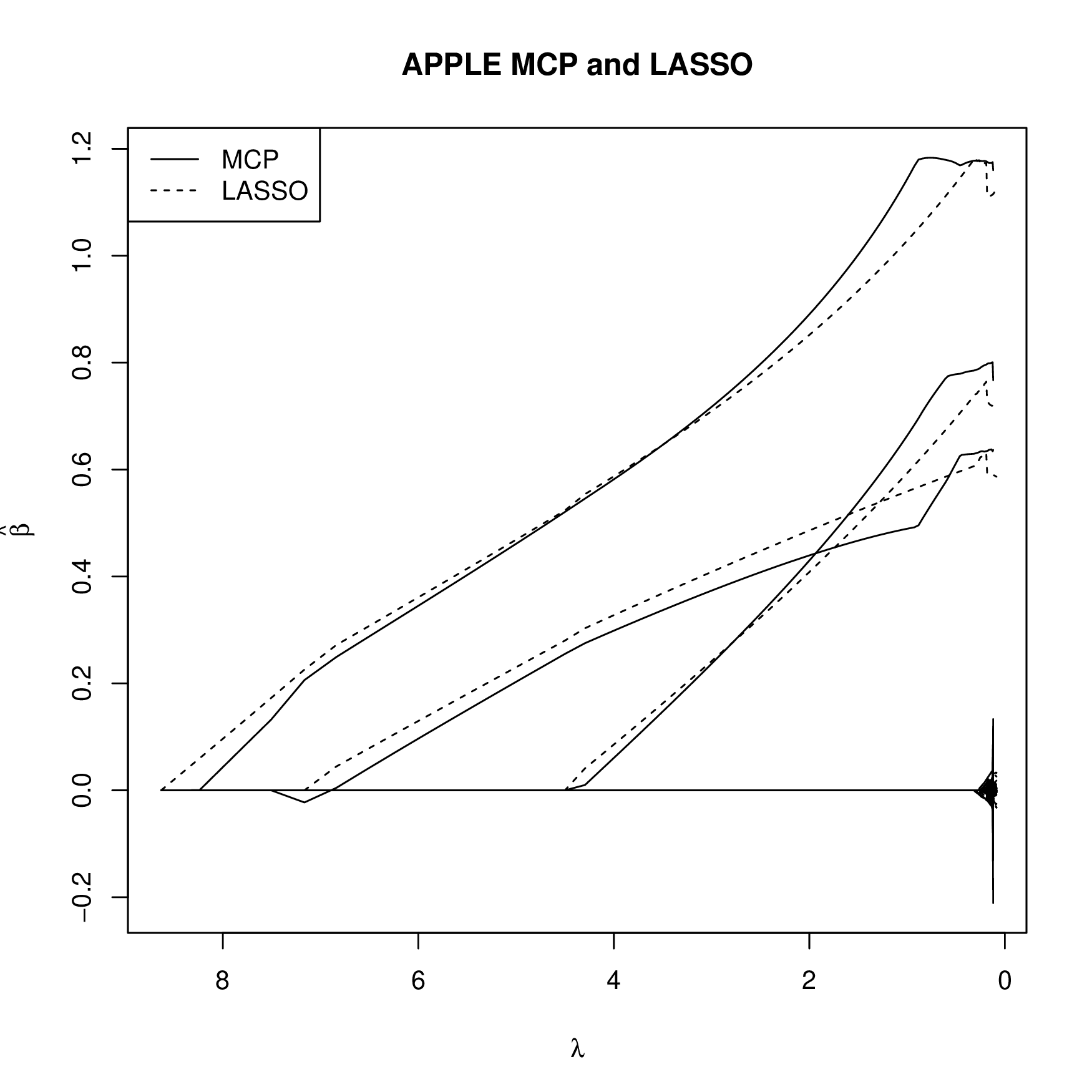}
\label{PL1}
}
\subfigure[]{
\includegraphics[scale=0.30]{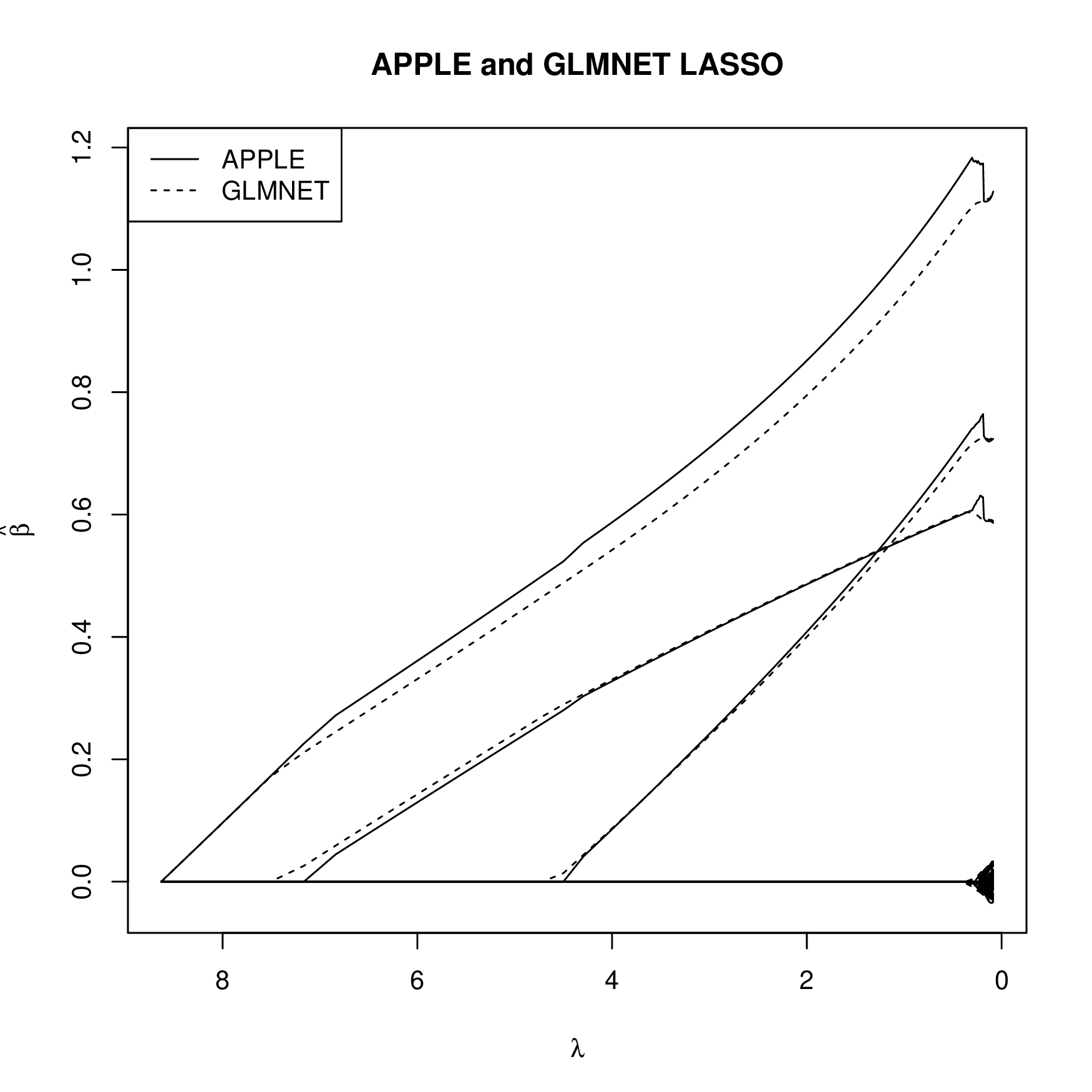}
\label{PL2}
}
\subfigure[]{
\includegraphics[scale=0.30]{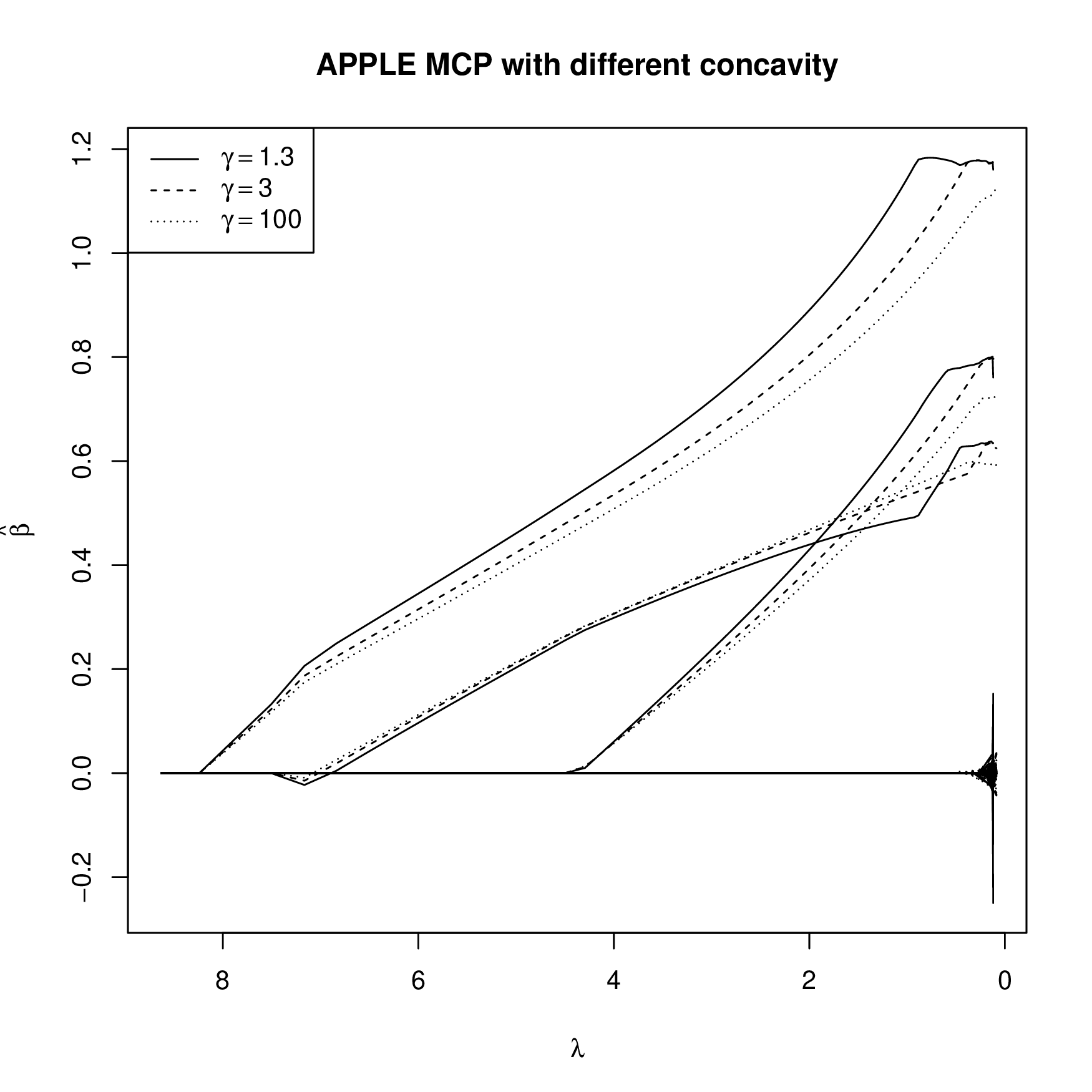}
\label{PL3}
}
\subfigure[]{
\includegraphics[scale=0.30]{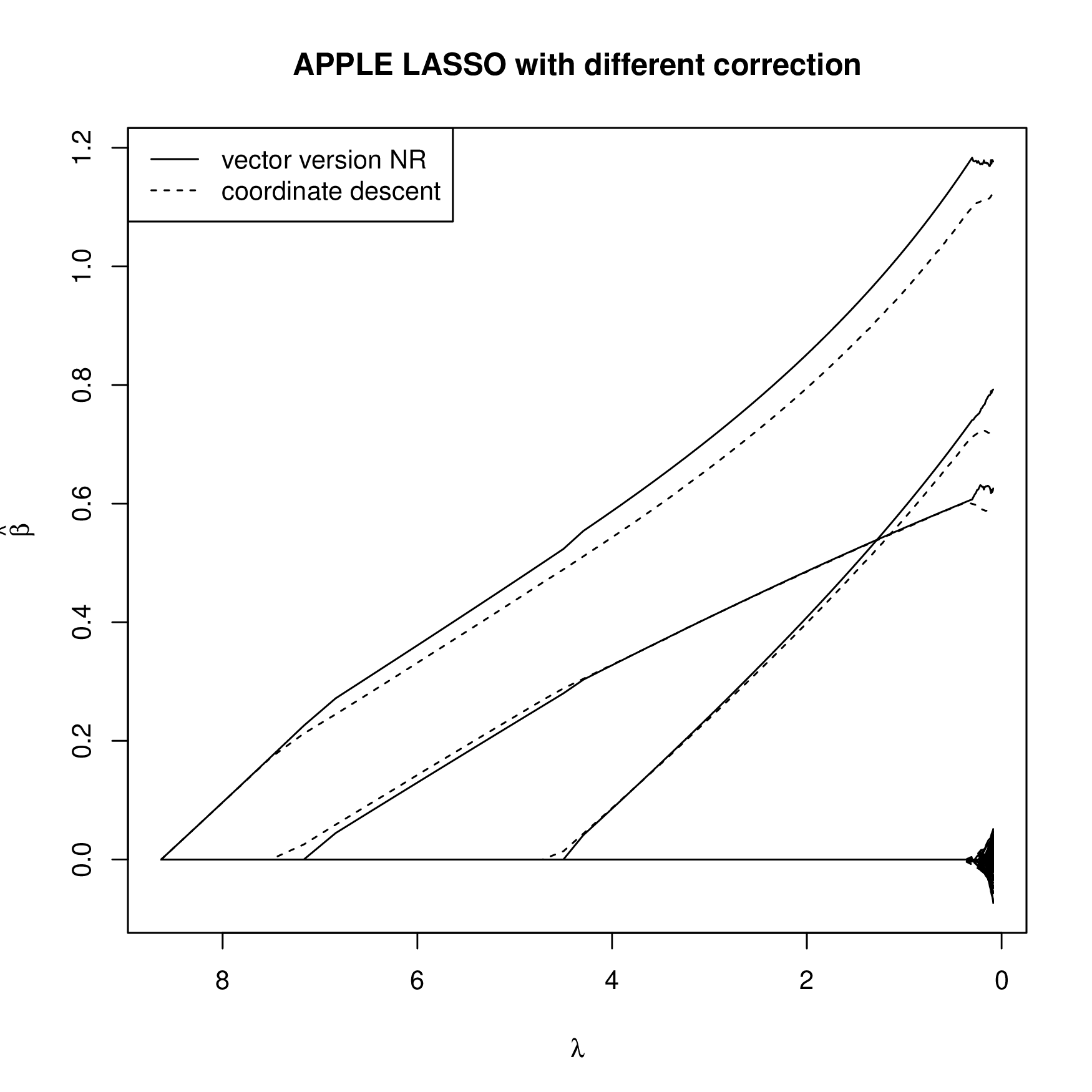}
\label{PL4}
}
\label{PE1}
\caption{Solution paths for poisson regression in Example \ref{PRE1}, where $p=1000$, $n=500$, $d=3$ and $\rho=0$. In (a), the solid lines and dotted lines are the solution paths for APPLE MCP and APPLE LASSO, respectively. In (b), the solid lines and dotted lines are the solution paths for APPLE LASSO and GLMNET LASSO, respectively. In (c), the solid lines, dashed lines and dotted lines are solution paths of APPLE MCP with different $\gamma$ values. In (d), the solid lines and dotted lines are the solution paths for APPLE LASSO with different correction methods. For each panel and each type of lines, the important variables are selected in the same order. As $\lam$ becomes smaller, variables with index 1, 5 and 2 are selected one by one. When $\lam$ gets near to zero, noise variables are selected.}
\end{figure*}

\begin{table*}
\caption{Comparison for APPLE with GLMNET for LASSO penalty and presenting APPLE MCP results, in Example \ref{PRE1}, where $p=1000$, $n=500$ and $d=3$. Design matrices with different correlation, and different selection criteria are presented. The medians of false positive (FP), true positive (TP), $\ell_1$ loss, and $\ell_2$ loss are reported over 100 repetitions, enclosed in parentheses are the corresponding standard errors.}\label{PET1}
\centering
\begin{tabular}[h]{l | l l |  c c c c}
\hline\noalign{\smallskip}
Model & Package & Method &FP & TP & $\ell_1$ loss & $\ell_2$ loss \\[0.5ex]
\noalign{\smallskip}\hline\noalign{\smallskip}
LASSO & APPLE & EBIC & 0.70(0.93) & 3.00(0.00) & 0.42(0.16) & 0.06(0.04)\\
$\rho=0$ & & CV & 9.41(3.75) & 3.00(0.00) & 0.33(0.11) & 0.02(0.01) \\
& GLMNET & EBIC & 0.66(1.20) & 3.00(0.00) & 0.89(0.15) & 0.24(0.08)\\
& & CV & 28.33(22.61) & 3.00(0.00) & 0.74(0.33) & 0.06(0.03) \\
\noalign{\smallskip}\hline\noalign{\smallskip}
MCP & APPLE & EBIC & 0.50(0.07) & 2.99(0.08) &  0.42(0.10)  & 0.06(0.03) \\
$\rho=0$, $\gamma=1.3$ & & CV &  2.89(2.76) & 2.99(0.26) & 0.49(0.27) & 0.07(0.05) \\
\noalign{\smallskip}\hline\noalign{\smallskip}
MCP & APPLE & EBIC  & 0.43(0.18) &  2.97(0.62) &  0.51(0.17) & 0.08(0.05) \\
$\rho=0$, $\gamma=3$ & & CV & 3.72(5.82) &  2.97(0.43) &  0.49(0.17) & 0.05(0.03) \\
\noalign{\smallskip}\hline\noalign{\smallskip}
LASSO & APPLE & EBIC & 0.56(0.80) & 3.00(0.00) & 0.37(0.13) & 0.05(0.03)\\
$\rho=0.2$ & & CV & 9.68(5.41) & 3.00(0.00) & 0.32(0.12) & 0.02(0.01)\\
& GLMNET & EBIC & 1.63(1.43) & 3.00(0.00) & 0.63(0.13) & 0.12(0.05)\\
& & CV & 55.56(25.21) & 3.00(0.00) & 2.61(0.67) & 0.08(0.03)\\
\noalign{\smallskip}\hline\noalign{\smallskip}
MCP & APPLE & EBIC &  0.81(2.39) & 2.97(0.16) & 0.35(1.19) & 0.19(1.16)\\
$\rho=0.2$, $\gamma=1.3$ & & CV & 6.60(15.19) & 2.97(0.16) & 1.82(12.20) & 4.85(14.29)\\
\noalign{\smallskip}\hline\noalign{\smallskip}
MCP & APPLE & EBIC  &  0.79(2.41) & 2.99(0.10) & 0.30(1.21) & 0.17(1.20)\\
$\rho=0.2$, $\gamma=3$ & & CV &  6.98(13.90) & 2.99(0.10) & 1.89(13.02) & 4.99(15.48) \\
\noalign{\smallskip}\hline\noalign{\smallskip}
LASSO & APPLE & EBIC & 0.86(0.95) & 3.00(0.00) & 0.29(0.11) & 0.03(0.03) \\
$\rho=0.5$ & & CV & 9.57(5.27) & 3.00(0.00) & 0.26(0.09) & 0.01(0.01) \\
& GLMNET & EBIC & 0.79(1.37) & 3.00(0.00) & 1.19(0.26) & 0.51(0.23)\\
& & CV & 25.64(13.92) & 3.00(0.00) & 0.54(0.17) & 0.04(0.02) \\
\noalign{\smallskip}\hline\noalign{\smallskip}MCP & APPLE & EBIC & 0.91(0.53) & 2.91(0.22) &  0.16(0.54)  & 0.00(0.45)\\
$\rho=0.5$, $\gamma=1.3$ & & CV & 6.12(2.07) & 2.97(0.28) & 0.28(0.78) & 0.03(0.75)  \\
\noalign{\smallskip}\hline\noalign{\smallskip}
MCP & APPLE & EBIC & 0.89(0.16) &  2.98(0.12) &  0.16(0.34) & 0.00(0.42) \\
$\rho=0.5$, $\gamma=3$ & & CV & 7.22(3.31) &  2.99(0.19) &  0.25(0.56) & 0.02(0.64) \\
\noalign{\smallskip}\hline\noalign{\smallskip}
LASSO & APPLE & EBIC & 1.01(0.93) & 3.00(0.00) & 0.18(0.13) & 0.01(0.01)\\
$\rho=0.7$ & & CV & 11.08(10.03) & 3.00(0.00) & 0.19(0.09) & 0.01(0.02)\\
& GLMNET & EBIC & 2.58(1.49) & 3.00(0.00) & 0.41(0.39) & 0.02(0.02)\\
& & CV & 27.39(18.42) & 3.00(0.00) & 0.43(0.20) & 0.06(0.05)\\
\noalign{\smallskip}\hline\noalign{\smallskip}
MCP & APPLE & EBIC & 1.02(0.19) & 2.98(0.12) & 0.18(0.10) & 0.01(0.01)\\
$\rho=0.7$, $\gamma=1.3$ & & CV & 16.10(6.10) & 3.00(0.00) & 0.15(0.11) & 0.00(0.01)\\
\noalign{\smallskip}\hline\noalign{\smallskip}
MCP & APPLE & EBIC  & 0.99(0.21) & 2.98(0.11) & 0.18(0.21) & 0.01(0.03)\\
$\rho=0.7$, $\gamma=3$ & & CV &  15.39(4.91) & 3.00(0.00) & 0.18(0.10) & 0.01(0.02)\\
\noalign{\smallskip}\hline
\end{tabular}
\end{table*}

\begin{table*}
\caption{Comparison for APPLE with GLMNET for LASSO penalty and presenting APPLE MCP results, in Example \ref{PRE1}, where $(p, n, d) = (3000, 500, 3)$ and $(1000, 500, 24)$. Design matrices with different correlation, and different selection criteria are presented. The medians of false positive (FP), true positive (TP), $\ell_1$ loss, and $\ell_2$ loss are reported over 100 repetitions, enclosed in parentheses are the corresponding standard errors. }
\label{PET2}
\centering
\begin{tabular}[h]{l | l l |  c c c c}
\hline\noalign{\smallskip}
Model & Package & Method &FP & TP & $\ell_1$ loss & $\ell_2$ loss \\[0.5ex]
\noalign{\smallskip}\hline\noalign{\smallskip}
LASSO & APPLE & EBIC & 0.49(0.25) & 3.00(0.00) & 0.33(0.10) & 0.02(0.02)\\
$p = 3000$ & & CV & 5.75(0.18) & 3.00(0.00) & 0.30(0.04)& 0.02(0.01)\\
$d = 3$ & GLMNET & EBIC & 1.10(0.13) & 3.00(0.00) & 0.56(0.23) & 0.06(0.02)\\
$\rho=0$& & CV &  98.28(12.17) & 3.00(0.00) & 1.36(0.32) & 0.04(0.20)\\
\noalign{\smallskip}\hline\noalign{\smallskip}
MCP & APPLE & EBIC &  1.38(0.47) & 3.00(0.00) & 0.29(0.04) & 0.02(0.01)\\
$p = 3000$, $d = 3$, $\rho = 0$ & & CV & 6.29(0.84) & 3.00(0.00) & 0.27(0.07) & 0.01(0.01) \\
\noalign{\smallskip}\hline\noalign{\smallskip}
LASSO & APPLE & EBIC & 0.67(0.18) & 3.00(0.00) & 0.29(0.04) & 0.03(0.01)\\
$p = 3000$ & & CV & 6.12(0.65) & 3.00(0.00) & 0.27(0.02) & 0.01(0.02)\\
$d = 3$ & GLMNET & EBIC & 1.92(0.54) & 3.00(0.00) & 0.64(0.01) & 0.12(0.01)\\
$\rho=0$& & CV & 168.53(6.76) & 3.00(0.00) & 2.13(0.02) & 0.12(0.02)\\
\noalign{\smallskip}\hline\noalign{\smallskip}
MCP & APPLE & EBIC & 2.39(0.55) & 3.00(0.00) & 0.26(0.02) & 0.02(0.01)\\
$p = 3000$, $d = 3$, $\rho = 0.5$ & & CV & 10.67(1.27) & 3.00(0.00) & 0.37(0.04) & 0.02(0.01) \\
\noalign{\smallskip}\hline\noalign{\smallskip}
LASSO & APPLE & EBIC & 0.73(0.18) & 15.29(0.47) & 3.92(0.53) & 2.88(0.23)\\
$p = 1000$ & & CV & 57.29(12.58) & 23.98(0.03) & 3.21(0.42) & 3.89(0.35)\\
$d = 24$ & GLMNET & EBIC & 0.65(0.10) & 0.49(0.04) & 4.28(1.20) & 3.26(0.63)\\
$\rho=0$& & CV & 142.48(22.58) & 23.97(0.10) & 4.29(1.02) & 5.32(0.37)\\
\noalign{\smallskip}\hline\noalign{\smallskip}
MCP & APPLE & EBIC &  0.27(0.01) & 22.19(0.27) & 1.02(0.02) & 0.98(0.02)\\
$p = 1000$, $d = 24$, $\rho = 0$ & & CV & 0.10(0.20) & 22.30(0.39) & 0.98(0.03) & 0.74(0.04)\\
\noalign{\smallskip}\hline\noalign{\smallskip}
LASSO & APPLE & EBIC & 0.70(0.23) & 12.14(2.49) & 4.29(1.02) & 5.20(0.43)\\
$p = 1000$ & & CV & 49.33(0.37) & 22.96(1.18) & 4.19(0.12) & 4.32(0.48)\\
$d = 24$ & GLMNET & EBIC & 0.22(0.20) & 1.02(0.04) & 5.21(0.53) & 4.94(0.39)\\
$\rho=0.5$& & CV &  155.32(13.28) & 23.01(0.20) & 6.32(0.48) & 5.23(0.47)\\
\noalign{\smallskip}\hline\noalign{\smallskip}
MCP & APPLE & EBIC &  0.25(0.10) & 22.47(1.03) & 1.20(0.07) & 0.99(0.17)\\
$p = 1000$, $d = 24$, $\rho = 0.5$ & & CV & 0.24(0.08) & 23.19(1.02) & 1.04(0.03) & 1.01(0.32)\\
\noalign{\smallskip}\hline
\end{tabular}
\end{table*}

In Figure \ref{PL1}, solution paths for APPLE MCP and LASSO are presented. Similar to the logistic regression case, LASSO yields a smoother path, while MCP results in better estimation with a nearly ``flat'' region of optimal level. In addition, at the end of the MCP path, the solution is still sparse in terms of our ``square root of sample size" criterion.
In Figure \ref{PL2}, we compare APPLE and GLMNET LASSO paths. As in the logistic regression model, there is a small jump in the APPLE LASSO path, which is caused by the change of correction method. After the correction method switches to the coordinate descent, the APPLE path coincides with the GLMNET LASSO path.
In Figure \ref{PL3}, APPLE MCP paths with different concavity parameters are presented. The continuous gradual change with respect to $\gamma$ is clear, with paths getting smoother and tending to the LASSO path as $\gamma$ becomes sufficiently large. In Figure \ref{PL4}, just as the logistic regression case, Newton-Raphson correction  yields more aggressive solution.
From the simulation results presented in Tables \ref{PET1} and \ref{PET2}, APPLE LASSO performs much better than GLMNET when CV is applied in all different $\rho$ cases. 
 Also, it is obvious that MCP does a better job than LASSO in terms of FP and TP. Similar behaviors as in logistic regression are observed when $d > \sqrt{n}$, as what we conjectured there, we think the main reason is the over-penalization of EBIC for the larger models.

Another interesting observation is the behavior when using different values of $\gamma$ for MCP in Figure \ref{PL3}, from the simulation results presented in Table \ref{PET1}, neither the selection nor the estimation seems to be sensitive to the choice of $\gamma$. This shows the stability of MCP in terms of the concavity parameter $\gamma$.

\subsection{Linear v.s. Quadratic Approximation}\label{sec::linear-vs-quadractic}

Different from most previous work where a linear approximation is used as the warm start in each update, APPLE uses a quadratic approximation. The technical details can be found in Sections \ref{subsec-lasso-update}, \ref{subsec-mcp-update} and Appendix. Here, we perform a simulation study to compare the solution path and the computation time under the settings $\rho=0$ and $\rho=0.5$ in Example \ref{BLE1}(i).

Due to the correction step used after the covariates are updated when $\lambda$ changes along the path, the linear and quadratic approximation yield essentially an identical solution path (not shown). The difference of these two methods lies in the quality of the warm starts in each step, which affects the computation cost. As expected, Table \ref{LvsQ} shows quadratic approximation saves time over linear approximation.

\begin{table*}
\caption{Comparison for linear and quadratic approximation in Section \ref{sec::linear-vs-quadractic}.  The median time (in seconds) are reported over 100 repetitions, enclosed in parentheses are the corresponding standard errors. CPU: Intel(R) Xeon(R) L5420 @ 2.50GHz. }
\label{LvsQ}
\centering
\begin{tabular}[h]{l l | c c | c c}
\hline\noalign{\smallskip}
& & \multicolumn{2}{c|}{LASSO} & \multicolumn{2}{c}{MCP} \\
& & $\rho = 0$ & $\rho = 0.5$ & $\rho = 0$ & $\rho = 0.5$ \\
\noalign{\smallskip}\hline\noalign{\smallskip}
 & Linear & 3.22(0.10) & 1.77(0.11) & 2.26(0.20)& 2.20(0.19)\\
& Quadratic & 1.50(0.12) & 1.56(0.14) & 1.90(0.20) & 2.02(0.20)\\
\noalign{\smallskip}\hline
\end{tabular}
\end{table*}

\section{Applications}\label{sec-application}

In this section, we present the analysis for two gene expression datasets with large dimension $p$ and small sample size $n$.

\begin{example}\label{AP1}
(i) We consider the leukemia dataset previously analyzed in \cite{Golub}. There are $p=7,129$ genes and $n=72$ samples coming from two classes: 47 in class ALL (acute lymphocytic leukemia) and 25 in class AML (acute myelogenous leukemia). (ii) The Neuroblastoma data set, obtained via the MicroArray Quality Control phase-II (MAQC-II) project \cite{MAQCII}, consists of gene expression profiles for $p = 10,707$ genes from 251 patients of the German Neuroblastoma Trials NB90-NB2004, diagnosed between 1989 and 2004. We analyzed the gene expression data with the 3-year event-free survival (3-year EFS), which indicates whether a patient survived 3 years after the diagnosis of neuroblastoma. There are $n = 239$ subjects with the 3-year EFS information available (49 positives and 190 negatives).
\end{example}

Potentially, a large number of genes are affected by the two types of leukemia in (i) or negative/positive information about 3-year EFS in (ii). In addition, the sample size $n$ is much smaller than the dimension $p$ for both problems. Therefore, a regularized logistic regression model is suitable. We impose LASSO and MCP penalties to these data sets, and compare the prediction accuracy yielded by the APPLE, GLMNET and NCVREG packages, respectively.

To check the stability of the results, we randomly split the data into training and testing sets 5 times for each example, and report the median prediction accuracy on the testing data and the median model size. For simplicity, EBIC was used to select the tuning parameter. For the MCP case, we fix $\gamma=1.3$ in both examples, which turned out to have better performance than larger $\gamma$ values in our simulation results. Notice that in some other high-dimensional variable selection literature,  a larger $\gamma$ was chosen to present the results. But when $\gamma$ is too large, the MCP solution path has little difference from the LASSO path, as shown in the figures of our simulation examples.

From the results in Table \ref{real}, where test error is the number of misclassified subjects out of the size of the test dataset, we notice that for LASSO, APPLE leads to  a smaller model size while having the same test error in both examples when compared with GLMNET. For MCP, in the leukemia example, APPLE only needs 1 variable to achieve the same test error as LASSO, and a better test error than the NCVREG. For the neuroblastoma example, MCP performs very well for both APPLE and NCVREG as compared with the LASSO.

\begin{table*}
\caption{Comparison for APPLE with GLMNET and NCVREG in LASSO and MCP ($\gamma=1.3$), respectively. The medians of model size and test error (the ratio of number of wrongly classified subjects to size of test dataset) for two real data sets are reported, enclosed in the enclosed in parentheses are the corresponding standard errors.}\label{real}
\centering
\begin{tabular}{l | l | l l | c c}
\hline\noalign{\smallskip}
Data & Criteria &\multicolumn{2}{c|}{LASSO}&\multicolumn{2}{c}{MCP ($\gamma=1.3$)} \\
& & APPLE & GLMNET & APPLE & NCVREG \\
\noalign{\smallskip}\hline\noalign{\smallskip}
leukemia & model size & 11 & 13 & 1 & 3 \\
& test error & 4/36 & 4/36 & 4/36 & 5/36\\
\noalign{\smallskip}\hline\noalign{\smallskip}
neuroblastoma & model size & 37 &  44 & 5 & 4 \\
& test error & 22/123 & 22/123 & 22/123 & 23/123\\
\noalign{\smallskip}\hline
\end{tabular}
\end{table*}

\section{Summary}\label{sec-summary}

In this paper, we propose a new algorithm, APPLE, for calculating the Approximate Path for Penalized Likelihood Estimators. The results from the simulation studies and real data examples provide compelling evidence that the APPLE algorithm is a worthwhile alternative to the existing methods.

APPLE takes significantly less time than NCVREG, and the same order of time as GLMNET. In each step, APPLE only updates the variables in the active set when the current model is sparse enough. When the model involves too many noise variables, APPLE switches to a coordinate descent correction.

The $\gamma$ adaptation method we adopt here is different from the one originally introduced by \cite{BrehenyHuang2011}. It is due to the vector update performed in APPLE. Here, the minimum eigenvalue adaptation  preserves the minimization of the maximum concavity of the MCP penalty while maintaining the stability in the Newton-Raphson update.

A public domain R language package \emph{apple} is available from the CRAN
website. \url{http://cran.r-project.org/web/packages/apple/}

\appendix


\section{Logistic Regression}

\subsection{LASSO}

In logistic regression, we assume $(\bx_i,~y_i)$, $i=1,\cdots, n$ are i.i.d. with $\mathbb{P}(y_i=1|\bx_i)=p_i=\exp(\bbeta'\bx_i)/(1+\exp(\bbeta'\bx_i))$. Then the target function for the LASSO penalized logistic regression is defined as
\[
L(\bbeta)=-\frac{1}{n}\sum_{i=1}^n \{y_i(\bbeta'\bx_i)-\log (1+\exp(\bbeta'\bx_i))\} +\lambda\sum_{j=1}^p |\beta_j|.
\]

The KKT conditions are given as follows.

\begin{align*}
  & \left\{
       \begin{array}{clcr}
            &\frac{1}{n}\sum_{i=1}^n\big\{\frac{\exp(\bhbeta'\bx_i)}{1+\exp(\bhbeta'\bx_i)}-y_i\big\}x_{ij}=\lambda \sgn(\hbeta_j),\quad &\hbeta_j\neq0;\\
            &|\frac{1}{n}\sum_{i=1}^n\big\{\frac{\exp(\bhbeta'\bx_i)}{1+\exp(\bhbeta'\bx_i)}-y_i\big\}x_{ij}|\le\lambda ,\quad &\hbeta_j=0;\\
            &\sum_{i=1}^n \frac{\exp(\bhbeta'\bx_i)}{1+\exp(\bhbeta'\bx_i)}= \sum_{i=1}^n y_i.
       \end{array}
   \right.
\end{align*}

We define active set $A_k$ as
\[
A_k=\bigg\{j:~ \big|\frac{1}{n}\sum_{i=1}^n \{y_i-\frac{\exp(\bhbeta^{(k)'} \bx_i)}{1+\exp(\bhbeta^{(k)'}\bx_i)}\}x_{ij}\big|\ge \lambda_k \bigg\} \cup \{0\}.
\]

To update, we define
\begin{align}
& \pi^{(k)}_i=\frac{\exp(\bhbeta^{(k)'}\bx_i)}{1+\exp(\bhbeta^{(k)'}\bx_i)}, \nonumber \\
& \bV^{(k)}=\text{diag}\{\pi^{(k)}_1(1-\pi^{(k)}_1), \cdots, \pi^{(k)}_n(1-\pi^{(k)}_n)\},\nonumber \\
& \bT^{(k)}=\text{diag}
\{\pi^{(k)}_1(1-\pi^{(k)}_1)\frac{1-\exp(\bhbeta^{(k)'}\bx_1)}{1+\exp(\bhbeta^{(k)'}\bx_1)},\cdots, \nonumber \\
& \pi^{(k)}_n(1-\pi^{(k)}_n)\frac{1-\exp(\bhbeta^{(k)'}\bx_n)}{1+\exp(\bhbeta^{(k)'}\bx_n)}\}, \label{T}
\end{align}
then $\bs^{(k)} = (0, \bs^{(k)'}_{-0})'$, $\bd^{(k)} = (0, \bd^{(k)'}_{-0})'$, where
\begin{align*}
& \bs^{(k)}_{-0}=-\bigg[\bX_{A_k\backslash\{0\}}^{'} \bV^{(k)}
 \bX_{A_k\backslash\{0\}}\bigg]^{-1}\sgn(\bhbeta^{(k)}_{A_k\backslash\{0\}}), \\
& \xi^{(k)}=\text{diag}(\bT^{(k)}\bX_{A_k\backslash\{0\}}\bs^{(k)}_{-0}), \\
& \bd^{(k)}_{-0}=-\bigg[ \bX_{A_k\backslash\{0\}}^{'} \bV^{(k)} \\
& \times \bX_{A_k\backslash\{0\}}\bigg]^{-1} \bX_{A_k\backslash\{0\}}^{'}\xi^{(k)}\bX_{A_k\backslash\{0\}} \bs^{(k)}_{-0}.
\end{align*}

To correct,
\begin{align*}
& \frac{\partial L^{(k)}}{\partial \bbeta_{A_k}}= \frac{1}{n}\bX_{A_k}^{'}\bigg(
\frac{\exp(\bhbeta^{(k)'}\bX)}{1+\exp(\bhbeta^{(k)'} \bX)}-y\bigg) \\
& +\lambda_k\sgn(0, \bhbeta^{(k)'}_{A_k\backslash\{0\}})', \\
& \frac{\partial^2 L^{(k)}}{\partial \bbeta_{A_k}\partial \bbeta_{A_k}^T}= \frac{1}{n} \bX_{A_k}^{'} \bV^{(k)} \bX_{A_k}.
\end{align*}


\subsection{MCP}

For MCP penalized logistic regression, we define the target function as
\begin{align*}
& L(\bbeta)=-\frac{1}{n}\sum_{i=1}^n \{y_i(\bbeta'\bx_i)-\log (1+\exp(\bbeta'\bx_i))\} \\
& +\lambda\sum_{j=1}^p \int_0^{|\beta_j|} (1-\frac{t}{\lambda\gamma})_+~dt.
\end{align*}

The KKT conditions are given as follows.
\begin{align*}
   \left\{
       \begin{array}{clcr}
            &\frac{1}{n}\sum_{i=1}^n\big\{\frac{\exp(\bhbeta'\bx_i)}{1+\exp(\bhbeta'\bx_i)}-y_i\big\}x_{ij}=\lambda (1-\frac{|\hbeta_j|}{\lambda\gamma}) \sgn(\hbeta_j),\\
             &\quad 0<|\hbeta_j|<\lambda\gamma.\\
            &\frac{1}{n}\sum_{i=1}^n\big\{\frac{\exp(\bhbeta'\bx_i)}{1+\exp(\bhbeta'\bx_i)}-y_i\big\}x_{ij}=0,\\
             &\quad |\hbeta_j|\ge\lambda\gamma.\\
            &|\frac{1}{n}\sum_{i=1}^n\big\{\frac{\exp(\bhbeta'\bx_i)}{1+\exp(\bhbeta'\bx_i)}-y_i\big\}x_{ij}|\le\lambda ,\\
             &\quad \hbeta_j=0.\\
            &\sum_{i=1}^n \frac{\exp(\bhbeta'\bx_i)}{1+\exp(\bhbeta'\bx_i)}= \sum_{i=1}^n y_i.
       \end{array}
   \right.\\
\end{align*}

For a given $\lambda_k$, define the active set $A_k$ as
\[
A_k=\{A_{k-1}\cup N_k\}\setminus D_k,
\]
where
\begin{align*}
& N_k=\{ j\in \{1,\cdots, p\}\setminus A_{k-1}: \\
& |\frac{1}{n}\sum_{i=1}^n\big\{\frac{\exp(\bhbeta'\bx_i)}{1+\exp(\bhbeta'\bx_i)}-y_i\big\}x_{ij} |\ge \lambda_k\},
\end{align*}
and
\begin{align*}
D_k=\left\{j\in A_{k-1}\cap A_{k-2}:~\sgn(\hbeta_j^{(k-1)})\sgn(\hbeta_j^{(k-2)})<0\right\}.
\end{align*}

To perform adaptive rescaling on $\gamma$, define
\begin{align*}
\boldsymbol{\Gamma}=\diag\left\{1/\gamma,\cdots,1/\gamma\right\}.
\end{align*}


To update, the derivatives are defined as follows,
\begin{align*}
& \bs^{(k)}_{-0}=-\bigg(\frac{1}{n}\bX_{A_k\backslash\{0\}}^{'}\bV^{(k)}
\bX_{A_k\backslash\{0\}}-u_{\text{min}}\boldsymbol{\Gamma}\bigg)^{-1}\\
& \times \sgn(\bhbeta^{(k)}_{A_k\backslash\{0\}}),\\
& \bd^{(k)}_{-0}=-\bigg[\frac{1}{n}\bX_{A_k\backslash\{0\}}^{'}\bV^{(k)}
\bX_{A_k\backslash\{0\}}-u_{\text{min}}\boldsymbol{\Gamma}\bigg]^{-1}\\
& \times \bX_{A_k\backslash\{0\}}^{'}\xi^{(k)}\bX_{A_k\backslash\{0\}}\bs^{(k)}_{-0},
\end{align*}
and
\begin{align*}
& \sgn(\bhbeta^{(k)}_{-0}) =
\begin{pmatrix}
\sgn(\hbeta_{A_{k,1}}^{(k)})I\{|\hbeta^{(k)}_{A_{k,1}}|<\lambda^{(k)}\gamma\} \\
\vdots\\
\sgn(\hbeta_{A_{k,n_k}}^{(k)})I\{|\hbeta^{(k)}_{A_{k, n_k}}|<\lambda^{(k)}\gamma\}\\
\end{pmatrix}, \\
& \sgn(\hbeta_j^{(k)})=-\frac{1}{n}\sum_{i=1}^n\big\{\frac{\exp(\bhbeta^{(k)'}\bx_i)}{1+\exp(\bhbeta^{(k)'}\bx_i)}-y_i\big\}x_{ij}.
\end{align*}

To correct we use
\[
\bhbeta^{(k,j+1)}_{A_k}=\bhbeta^{(k,j)}_{A_k}-\bigg(\frac{\partial^2 L^{(k)}}{\partial \bbeta_{A_k}\partial \bbeta_{A_k}^T}\bigg)^{-1} \bigg(\frac{\partial L^{(k)}}{\partial \bbeta_{A_k}}\bigg),
\]
where
\begin{align}
& \frac{\partial L^{(k)}}{\partial \bbeta_{A_k}}=-\frac{1}{n}\bX_{A_k}^{'}\bV^{(k)}\bigg({\bf \tilde{y}}-\bX_{A_k}\bhbeta^{(k)}_{A_k}\bigg) \nonumber\\
& +\lambda_k\sgn(0, \bhbeta^{(k)'}_{A_k\backslash\{0\}})'\big(1-\frac{|\bhbeta^{(k)}_{A_k\backslash\{0\}}|}{\lambda_k\gamma}\big)_+,\label{piV11}\\
& \frac{\partial^2 L^{(k)}}{\partial \bbeta_{A_k}\partial \bbeta_{A_k}^T} = \frac{1}{n} \bX_{A_k}^{'} \bV^{(k)} \bX_{A_k}-u_{\text{min}}\boldsymbol{\Gamma}, \label{piV12}
\end{align}
and
\begin{align}
\tilde{\by}=\left(\bV^{(k)}\right)^{-1}\left\{\by - \frac{\exp\left(\bhbeta^{(k)'}\bX\right)}{1+\exp\left(\bhbeta^{(k)'}\bX\right)}\right\}.\label{piV13}
\end{align}

\section{Poisson Regression}

\subsection{LASSO}

In Poisson regression, we assume $(\bx_i,~y_i),~i=1,\cdots, n$ are iid with $\mathbb{P}(Y=y_i)=e^{-\lambda_i}\lambda_i^{y_i}/(y_i)!$, where $\log \lambda_i=\bbeta'\bx_i$. Then criterion for the LASSO penalized Poisson regression is defined as
\[
L(\bbeta)=\frac{1}{n}\sum_{i=1}^n \big\{\exp(\bbeta'\bx_i)-y_i(\bbeta'\bx_i)\big\}+\lambda\sum_{j=1}^p |\beta_j|.
\]

The KKT conditions are given as follows.
\begin{align*}
   \left\{
       \begin{array}{clcr}
            &\frac{1}{n}\sum_{i=1}^n\big\{\exp(\bhbeta'\bx_i)-y_i\big\}x_{ij}=\lambda \sgn(\hbeta_j),\quad &\hbeta_j\neq0.\\
            &|\frac{1}{n}\sum_{i=1}^n\big\{\exp(\bhbeta'\bx_i)-y_i\big\}x_{ij}|\le\lambda ,\quad &\hbeta_j=0.\\
            &\hbeta_0=\log \frac{\sum_{i=1}^n y_i}{\sum_{i=1}^n \exp(\bhbeta'\bx_i)}.
       \end{array}
   \right.
\end{align*}

For a given $\lambda_k$, we define the active set $A_k$ as follows.
\[
A_k=\big\{j:~ \big|\frac{1}{n}\sum_{i=1}^n \{y_i-\exp(\bhbeta^{(k)'}\bx_i)x_{ij}\}\big|\ge \lambda_k\big\}\cup \{0\}.
\]

To update, we define
\begin{align*}
\bV^{(k)}=\text{diag}\{\exp(\bhbeta^{(k)'}\bx_1), \cdots,\exp(\bhbeta^{(k)'}\bx_n)\},
\end{align*}
then
\begin{align*}
& \bs^{(k)}_{-0}=-\bigg[ \bX_{A_k\backslash\{0\}}^{'} \bV^{(k)}
 \bX_{A_k\backslash\{0\}}\bigg]^{-1}\sgn(\bhbeta^{(k)}_{A_k\backslash\{0\}}), \\
& \boldsymbol{\xi}^{(k)}=\text{diag}(\bV^{(k)}\bX_{A_k\backslash\{0\}}\bs^{(k)}_{-0}),
\end{align*}
and
\begin{align*}
& \bd^{(k)}_{-0}=-\bigg[ \bX_{A_k\backslash\{0\}}^{'} \bV^{(k)}  \bX_{A_k\backslash\{0\}}\bigg]^{-1} \\
&\times \bX_{A_k\backslash\{0\}}^{'}\boldsymbol{\xi}^{(k)}\bX_{A_k\backslash\{0\}} \bs^{(k)}_{-0}.
\end{align*}

To correct,
\begin{align*}
& \frac{\partial L^{(k)}}{\partial \bbeta_{A_k}}= \frac{1}{n}\bX_{A_k}^{'}\bigg(\exp(\bhbeta^{(k)'}\bX)-y\bigg) +\lambda_k\sgn(0, \bhbeta^{(k)'}_{A_k\backslash\{0\}})', \\
& \frac{\partial^2 L^{(k)}}{\partial \bbeta_{A_k}\partial \bbeta_{A_k}^T}= \frac{1}{n} \bX_{A_k}^{'} \bV^{(k)} \bX_{A_k}.
\end{align*}


\subsection{MCP}

For MCP penalized Poisson regression, we define the target function as
\begin{align*}
& L(\bbeta)=\frac{1}{n}\sum_{i=1}^n \big\{\exp(\bbeta'\bx_i)-y_i(\bbeta'\bx_i)\big\} \\
& +\lambda\sum_{j=1}^p \int_0^{|\beta_j|} (1-\frac{t}{\lambda\gamma})_+~dt.
\end{align*}

The KKT conditions are,
\begin{align*}
    \left\{
       \begin{array}{clcr}
            &\frac{1}{n}\sum_{i=1}^n\big\{\exp(\bhbeta'\bx_i)-y_i\big\}\bx_{ij}=\lambda (1-\frac{|\bhbeta_j|}{\lambda\gamma}) \sgn(\hbeta_j), \\
            &\quad 0<|\hbeta_j|<\lambda\gamma.\\
            &\frac{1}{n}\sum_{i=1}^n\big\{\exp(\bhbeta'\bx_i)-y_i\big\}\bx_{ij}=0, \\
            & \quad |\hbeta_j|\ge\lambda\gamma.\\
            &|\frac{1}{n}\sum_{i=1}^n\big\{\exp(\bhbeta'\bx_i)-y_i\big\}\bx_{ij}|\le\lambda ,\\
            & \quad \hbeta_j=0.\\
            &\hbeta_0=\log \frac{\sum_{i=1}^n y_i}{\sum_{i=1}^n \exp(\bhbeta'\bx_i)}.
       \end{array}
   \right.
\end{align*}

For a given $\lambda_k$, the active set is defined as
\[
A_k=\{A_{k-1}\cup N_k\}\setminus D_k,
\]
where
\begin{align*}
 & N_k = \big\{ j\in \{1,\cdots, p\}\setminus A_{k-1}:\\
 &|\frac{1}{n}\sum_{i=1}^n \{\exp(\bhbeta'\bx_i)-y_i\}\bx_{ij}|\ge \lambda_k\big\},
\end{align*}
and
\begin{align*}
D_k = \big\{j\in A_{k-1}\cap A_{k-2}:~\sgn(\hbeta_j^{(k-1)})\sgn(\hbeta_j^{(k-2)})<0\big\}.
\end{align*}

To update, the derivatives are defined as follows,
\begin{align*}
& \bs^{(k)}_{-0}=\bigg(\frac{1}{n}\bX_{A_k\backslash\{0\}}^{'}\bV^{(k)}
\bX_{A_k\backslash\{0\}}-u_{\text{min}}\boldsymbol{\Gamma}\bigg)^{-1}\\
&\times\sgn(\bhbeta^{(k)}_{A_k\backslash\{0\}}),\\
& \bd^{(k)}_{-0}=-\bigg[\frac{1}{n}\bX_{A_k\backslash\{0\}}^{'}\bV^{(k)} \bX_{A_k\backslash\{0\}}-u_{\text{min}}\boldsymbol{\Gamma}\bigg]^{-1}\\
&\times \bX_{A_k\backslash\{0\}}^{'}\xi^{(k)} \bX_{A_k\backslash\{0\}}\bs^{(k)}_{-0},
\end{align*}
and
\begin{align*}
& \sgn(\bhbeta^{(k)}_{-0})=
\begin{pmatrix}
\sgn(\bhbeta_{A_{k,1}}^{(k)})I\{|\bhbeta^{(k)}_{A_{k,1}}|<\lambda^{(k)}\gamma\} \\
\vdots\\
\sgn(\bhbeta_{A_{k, n_k}}^{(k)})I\{|\bhbeta^{(k)}_{A_{k, n_k}}|<\lambda^{(k)}\gamma\}\\
\end{pmatrix}, \\
& \sgn(\hbeta_j^{(k)})=-\frac{1}{n}\sum_{i=1}^n\big\{\exp(\bhbeta^{(k)'}\bx_i)-y_i\big\}x_{ij}.
\end{align*}

To correct we use
\begin{align*}
\bhbeta^{(k,j+1)}_{A_k}=\bhbeta^{(k,j)}_{A_k}-\bigg(\frac{\partial^2 L^{(k)}}{\partial \bbeta_{A_k}\partial \bbeta_{A_k}^T}\bigg)^{-1} \bigg(\frac{\partial L^{(k)}}{\partial \bbeta_{A_k}}\bigg),
\end{align*}
where
\begin{align}
&\frac{\partial L^{(k)}}{\partial \bbeta_{A_k}}=-\frac{1}{n}\bX_{A_k}^{'}V^{(k)}\bigg({\bf \tilde{y}}-\bX_{A_k}\bhbeta^{(k)}_{A_k}\bigg) \nonumber\\
& +\lambda_k\sgn(0, \bhbeta^{(k)'}_{A_k\backslash\{0\}})'\big(1-\frac{|\bhbeta^{(k)}_{A_k\backslash\{0\}}|}{\lambda_k\gamma}\big)_+,\label{piV21}\\
&\frac{\partial^2 L^{(k)}}{\partial \bbeta_{A_k}\partial \bbeta_{A_k}^T}= \frac{1}{n} \bX_{A_k}^{'} \bV^{(k)} \bX_{A_k}-u_{\text{min}}\boldsymbol{\Gamma}, \label{piV22}
\end{align}
and
\begin{align}
\tilde{\by}=\left(\bV^{(k)}\right)^{-1}\left\{\by-\exp\left(\bhbeta^{(k)'}\bX\right)\right\}.\label{piV23}
\end{align}

\section*{Acknowledgments}
The authors thank the editor, the associate editor, and referees for their constructive comments.
The authors thank Diego Franco Salda\~na for proofreading.

\bibliographystyle{spbasic}
\bibliography{VariSele}

\end{document}